\documentclass[journal]{IEEEtran}
\usepackage[pdftex]{graphicx}
\usepackage{url}
\usepackage[numbers,sort,compress]{natbib}
\usepackage{amsmath}
\usepackage{amssymb}
\usepackage{adjustbox}
\usepackage{multirow}
\usepackage{algorithm2e}
\usepackage{hyperref}
\usepackage{cleveref}
\usepackage{pgfplots}
\Crefname{algocf}{Algorithm}{Algorithms}
\usepackage{textcomp}
\usepackage{bbm}
\usepackage{tikz}
\usepackage{booktabs}
\usepackage{amstext} 
\usepackage{array}   
\newcolumntype{L}{>{$}c<{$}} 

\usepackage{caption}
\usepackage[flushleft]{threeparttable}
\usepackage{pifont}

\usepackage[caption=false, font=footnotesize]{subfig}
\usepackage{subcaption}
\usepackage{lscape}
\usepackage{soul}
\usepackage{color,colortbl}
\usepackage{multicol}
\usepackage{numprint}
\usepackage{url}
\hypersetup{colorlinks=true}

\newlength{\clinegapskip}
\newcommand{\midsepremove}{\aboverulesep = 0mm \belowrulesep = 0mm}
\midsepremove
\newcommand{\midsepdefault}{\aboverulesep = 0.605mm \belowrulesep = 0.984mm}
\midsepdefault

\definecolor{bareland}{rgb}{0.50196078,0,0}
\definecolor{rangeland}{rgb}{0,1,0.14117647}
\definecolor{develop}{rgb}{0.58039216,0.58039216,0.58039216}
\definecolor{road}{rgb}{1,1,1}
\definecolor{tree}{rgb}{0.13333333,0.38039216,0.14901961}
\definecolor{water}{rgb}{0,0.27058824,1}
\definecolor{agriculture}{rgb}{0.29411765,0.70980392,0.28627451}
\definecolor{building}{rgb}{0.87058824,0.12156863,0.02745098}

\usepackage{balance}

\pgfplotsset{compat=1.18} 

\begin{document}
\title{Real-Time Semantic Segmentation: A Brief Survey \& Comparative Study in Remote Sensing}

\author{Clifford Broni-Bediako,~\IEEEmembership{Member,~IEEE,}
Junshi~Xia,~\IEEEmembership{Senior Member,~IEEE,}
        Naoto~Yokoya,~\IEEEmembership{Member,~IEEE}
\thanks{Manuscript received; revised. This work was supported by KAKENHI, Grant Number 22H03609.}
\thanks{C. Broni-Bediako and J. Xia are with Geoinformatics Team, RIKEN Center for Advanced Intelligence Project (AIP), Japan (email: clifford.broni-bediako@riken.jp and junshi.xia@riken.jp). 
\par
N. Yokoya is with the Department of Complexity Science and Engineering, The University of Tokyo, and Geoinformatics Team, RIKEN Center for Advanced Intelligence Project (AIP), Japan (email: yokoya@k.u-tokyo.ac.jp).
}
}

\maketitle

\begin{abstract}
Real-time semantic segmentation of remote sensing imagery is a challenging task that requires a trade-off between effectiveness and efficiency. It has many applications including tracking forest fires, detecting changes in land use and land cover, crop health monitoring, and so on. With the success of efficient deep learning methods (i.e., efficient deep neural networks) for real-time semantic segmentation in computer vision, researchers have adopted these efficient deep neural networks in remote sensing image analysis. This paper begins with a summary of the fundamental compression methods for designing efficient deep neural networks and provides a brief but comprehensive survey, outlining the recent developments in real-time semantic segmentation of remote sensing imagery. We examine several seminal efficient deep learning methods, placing them in a taxonomy based on the network architecture design approach. Furthermore, we evaluate the quality and efficiency of some existing efficient deep neural networks on a publicly available remote sensing semantic segmentation benchmark dataset, the OpenEarthMap. The experimental results of an extensive comparative study demonstrate that most of the existing efficient deep neural networks have good segmentation quality, but they suffer low inference speed (i.e., high latency rate), which may limit their capability of deployment in real-time applications of remote sensing image segmentation. We provide some insights into the current trend and future research directions for real-time semantic segmentation of remote sensing imagery.
\end{abstract}
		
\begin{IEEEkeywords}
real-time semantic segmentation, efficient deep neural networks, remote sensing image analysis
\end{IEEEkeywords}
		
\IEEEpeerreviewmaketitle

\section{Introduction}\label{sec:1}
Semantic segmentation is a problem of labelling each pixel in an image with a class label to partition the image into semantically meaningful segments. It is a pixel-level classification problem as compared to image classification which assigns a class label to the entire image \cite{szeliski2022}. In recent years, efficient deep neural networks (DNNs) for real-time semantic segmentation have held a great position in the computer vision community \cite{li2019dfanet,liu2019FeaturePE,li2020semantic,yang2021ddpn,orsic2019}.  Efficient DNNs are neural networks with low computational footprints and inference time \cite{sze2020efficient}. The success of these methods has made a considerable impact in real-time applications in the fields of autonomous vehicles \cite{wenfu2020,papadeas2021}, robot vision \cite{mahe2019,bruce2000}, and medical image analysis \cite{xiaogang2022,lou2023} which require high-end computer vision systems. This promising progress of the efficient DNNs in real-time applications of image semantic segmentation has sparked an interest in this topic in the remote sensing community \cite{rs14236057,land10010079}. Because many real-world applications of remote sensing image semantic segmentation such as flood detection \cite{9939092}, burned area detection \cite{9852471}, monitoring of weeds in farmlands \cite{app10207132}, and so on are required to operate in real time and more appropriately on resource-constrained devices. Nevertheless, most of the state-of-the-art DNNs for semantic segmentation of remote sensing imagery require high-powered general-purpose machines (e.g., GPUs) \cite{yuan2021,jiang2022}, which greatly limits their applications in real time in a resource-constrained environment. Thus, it is important to improve the design of DNNs to achieve efficient networks that can enable the development of remote sensing image semantic segmentation methods for real-time applications. 

\begin{figure}
    \centering
    \includegraphics[width=1\linewidth, height=0.65\linewidth]{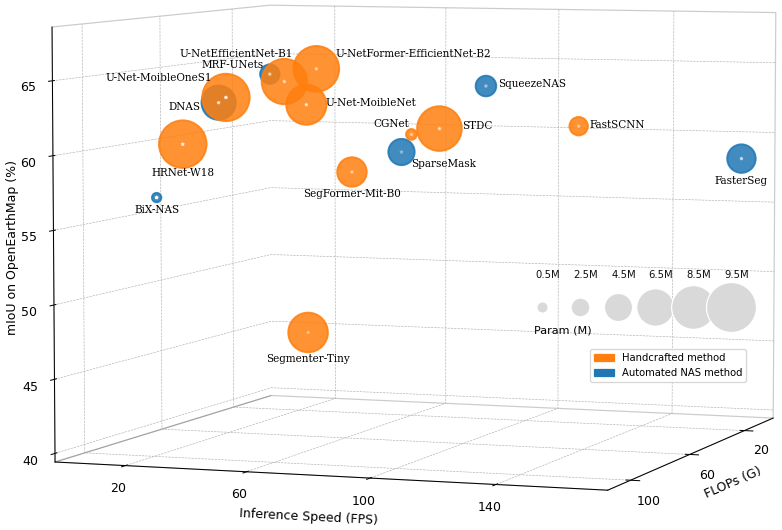}
    \captionsetup{width=1\linewidth}
    \caption{The performance comparison of some existing efficient deep neural networks for real-time semantic segmentation on the OpenEarthMap benchmark. The bubble size denotes the number of parameters (Param). The FLOPs and FPS (frames per second) were computed on 1024x1024 RGB input data. The inference speed was computed on a single NVIDIA Tesla P100 (DGX-1) with 16 GB memory based on an average run-time of 300 iterations with 10 iterations warm-up.}
    \label{fig:comp-iou-fps}
    \vspace{-5mm}
\end{figure}

In the literature, most of the surveys and reviews published on image semantic segmentation in the remote sensing community \cite{yuan2021,bipul2021,jiang2022} (or in the computer vision \cite{hao2020,mo2022,minaee2022,ulku2022}) are about deep learning methods in general. Yuan \textit{et al.}~\cite{yuan2021} and Jiang \textit{et al.}~\cite{jiang2022} reviewed deep learning methods (such as fully convolutional networks \cite{fcnChen2018,fcnHan2020}, feature pyramid networks \cite{Venugopal2020}, encoder-decoder networks \cite{unetCui2019,unetCheng2020}, etc.) for semantic segmentation of remote sensing imagery. Neupane \textit{et al.}~\cite{bipul20211} performed a meta-analysis to review and analyze DNNs-based semantic segmentation methods in remote sensing applications. In addition, Guo \textit{et al.}~\cite{Guo2018} summarized DNNs methods for semantic segmentation of remote sensing optical imagery and categorized them into region-based, fully convolutional-based and weakly supervised semantic segmentation methods. Recently, Gama \textit{et al.}~\cite{GAMA2023} and Catalano \textit{et al.}~\cite{catalano2023shot} have reviewed meta-learning and few-shot learning methods \cite{lang2022learning,lang2023,langcheng2023,langchunbo2023} for semantic segmentation. Very little work has sought to examine the efficient DNNs methods for real-time semantic segmentation \cite{holder2022,papadeas2021,takos2020}. The work in \cite{holder2022} and \cite{takos2020} reviewed real-time semantic segmentation in computer vision in general, while \cite{papadeas2021} examined the design of efficient DNNs for real-time semantic segmentation in the field of autonomous vehicles. Remote sensing applications require specialized algorithms and techniques to process large volumes of data collected by sensors capturing the earth's surface images at different wavelengths. Variability in environmental conditions poses a challenge for extracting meaningful information from the data. Integrating data from multiple sources can also be challenging, making remote sensing applications unique compared to other fields like autonomous driving. To the best of our knowledge, no work in the literature has examined the efficient DNNs methods for real-time semantic segmentation towards real-time applications in remote sensing image understanding. To this end, this paper aims to address this gap by summarising most of the state-of-the-art efficient DNNs methods in the literature that have been instrumental in real-time semantic segmentation of remote sensing imagery. Furthermore, to enable practitioners and researchers in the remote sensing community to adopt the best efficient DNNs for real-time applications of remote sensing image semantic segmentation, we present a comparative study of several established efficient deep neural networks for real-time semantic segmentation on OpenEarthMap \cite{Xia2023WACV} remote sensing image semantic segmentation benchmark. Fig.~\ref{fig:comp-iou-fps} shows the performance in terms of accuracy (mIoU), inference speed (FPS), the number of floating-point operations (FLOPs), and the number of parameters (Params) of the models used in the comparative study on the OpenEarthMap benchmark. 

The closely related work is the comparative study conducted by Safavi and Rahnemoonfar \cite{9939092}, which evaluated several efficient DNNs methods for real-time semantic segmentation on the FloodNet dataset \cite{radosavovic2020designing} of aerial imagery. The authors provided a detailed analysis of the segmentation accuracy and efficiency of the methods in a real-time post-disaster assessment of remote sensing. The contribution of this paper is different from \cite{9939092} in the following perspectives:
\begin{itemize}
    \item[1.] This paper presents a comprehensive summary of compression techniques and efficiency metrics that are commonly used in designing efficient deep neural networks for real-time semantic segmentation.
    \item[2.] It summarizes the most seminal literature on efficient deep learning methods for real-time semantic segmentation of remote sensing imagery.
    \item[3.] Also, it provides a comparative study (in terms of quality-cost trade-off) of some of the existing efficient deep neural networks (not only handcrafted ones as in \cite{9939092}, but includes automated architecture-searched networks) for real-time semantic segmentation. 
    \item[4.] The study was also extended to investigate continent-wise domain generalisation semantic segmentation of the networks using the continent-wise domain adaptation settings in \cite{Xia2023WACV}. Here, both handcrafted and automated architecture-searched networks are compared as well.
    \item[5.] Finally, it presents an insightful discussion on challenges in real-time semantic segmentation of remote sensing imagery. 
\end{itemize}

The rest of this paper is organized as follows. Section \ref{sec:2} briefly summarizes the design approaches of efficient DNNs. The most seminal models are summarized in Section \ref{sec:3}. In Section \ref{sec:4}, we present the settings for the comparative study of representative models on the OpenEarthMap benchmark. Then, the discussion of the findings of the study is presented in Section \ref{sec:5}. Conclusion and some open challenges for a future investigation are made in Section \ref{sec:5}. 

\section{Efficient DNNs Design Approach}\label{sec:2}
There has been a great deal of research in recent years on efficient DNNs methods for real-time applications and resource-constrained platforms \cite{ray2022tiny,sze2017,nan2019comp}. Here, we provide a brief but comprehensive overview of model compression techniques and efficiency metrics that have been proposed and extensively used in developing real-time semantic segmentation networks. 

\subsection{Compression Techniques}\label{sec:2.2}
As with any task deep learning algorithm is employed for, once the algorithm is trained, it is used to perform the task by using its weights learned during training to compute the output. This is referred to as \textit{inference}. In the particular case of semantic segmentation tasks in real time, besides the output quality, the primary concern is efficient inference, which involves optimising the architecture of a model by reducing its computational complexity and memory footprint to speed up inference \cite{menghani2023}. Several compression techniques \cite{lecun1989} have been proposed to optimize an architecture of a model for one or more of the aforementioned efficiency metrics: number of parameters, memory consumption, and latency, to achieve efficient inference with a trade-off of the model's quality \cite{menghani2023}. Here, we briefly summarize the commonly used model compression techniques and refer readers to \cite{deng2020model,choudhary2020,berthelier2020} for a more detailed discussion on this subject.

\begin{figure*}
\centering
    \subfloat[]{\includegraphics[width=0.15\linewidth, height=0.1\linewidth]{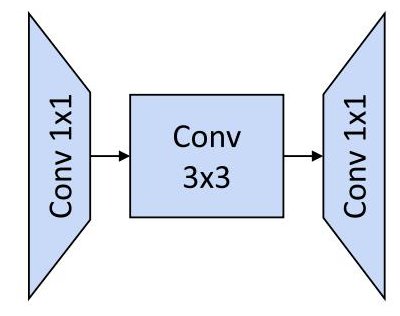}}\quad
    \subfloat[]{\includegraphics[width=0.15\linewidth, height=0.1\linewidth]{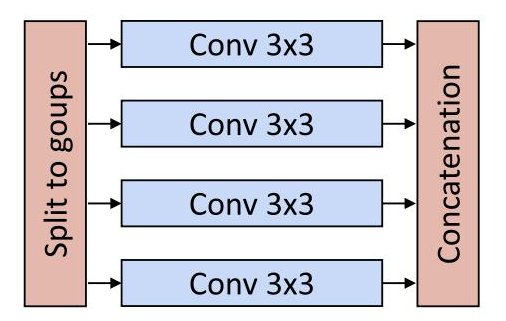}}\quad
    \subfloat[]{\includegraphics[width=0.15\linewidth, height=0.1\linewidth]{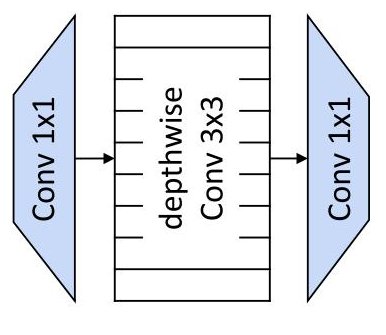}}\quad
    \subfloat[]{\includegraphics[width=0.15\linewidth, height=0.1\linewidth]{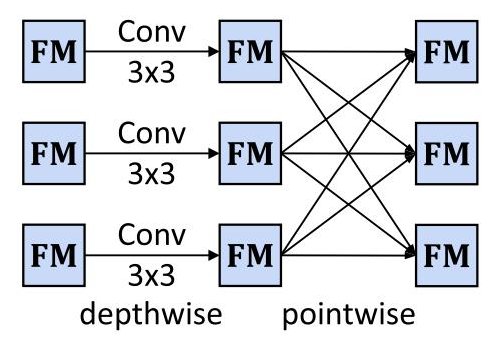}}\\

    \subfloat[]{\includegraphics[width=0.3\linewidth, height=0.15\linewidth]{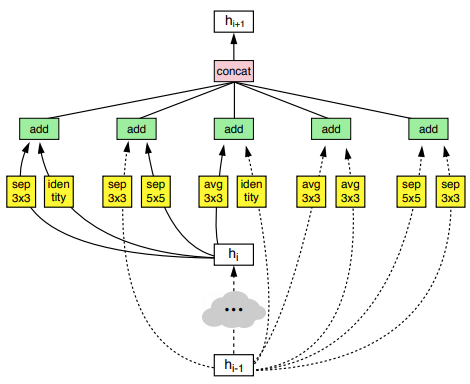}}\qquad
    \subfloat[]{\includegraphics[width=0.3\linewidth, height=0.15\linewidth]{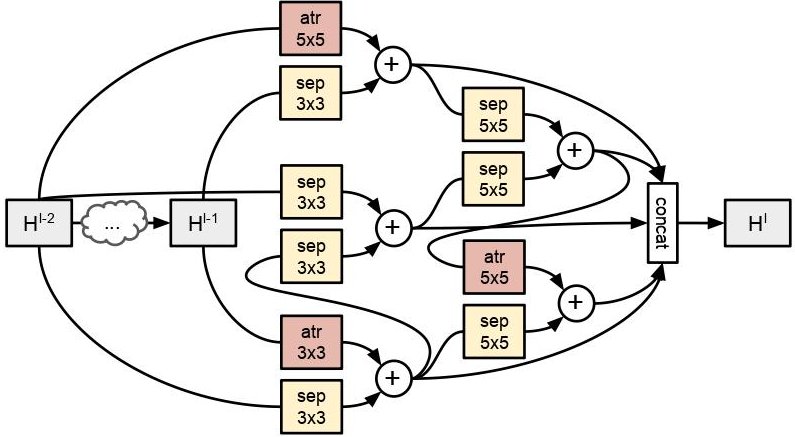}}\\

    \subfloat[]{\includegraphics[width=0.13\linewidth, height=0.13\linewidth]{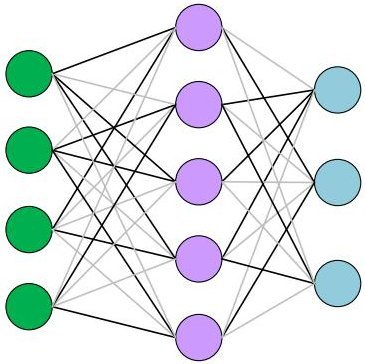}}\quad
    \subfloat[]{\includegraphics[width=0.13\linewidth, height=0.13\linewidth]{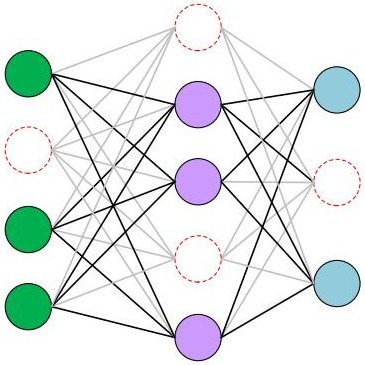}}\qquad
    \subfloat[]{\raisebox{-0.3\height}{\includegraphics[width=0.33\linewidth, height=0.27\linewidth]{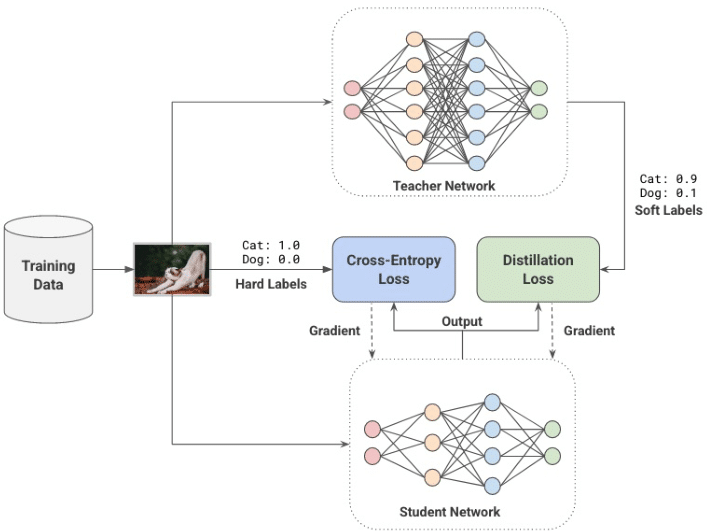}}}\\
    
    \subfloat[]{\includegraphics[width=0.128\linewidth, height=0.128\linewidth]{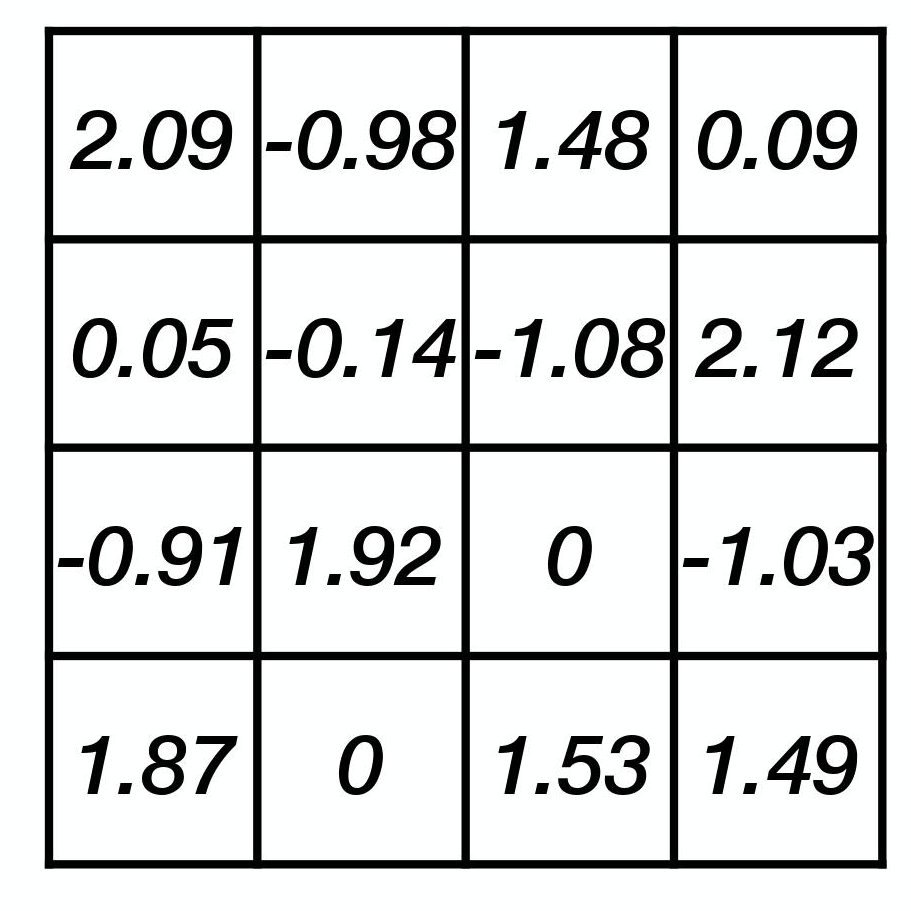}}\quad
    \subfloat[]{\includegraphics[width=0.128\linewidth, height=0.128\linewidth]{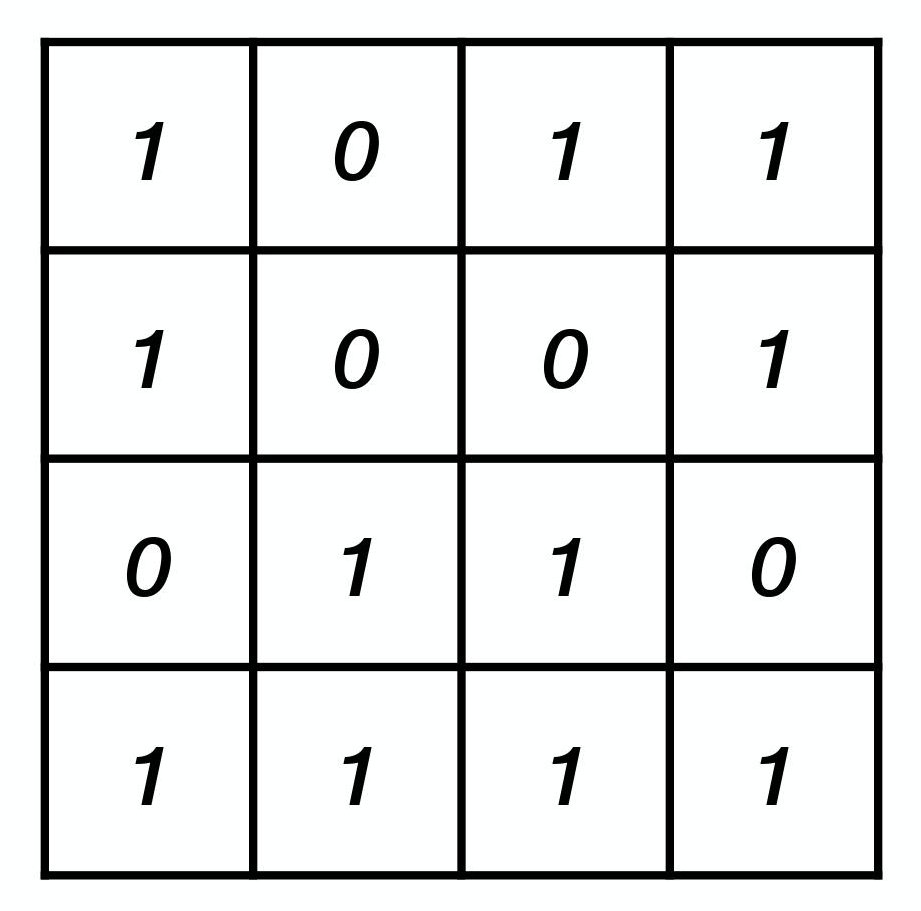}}\quad
    \subfloat[]{\includegraphics[width=0.18\linewidth, height=0.128\linewidth]{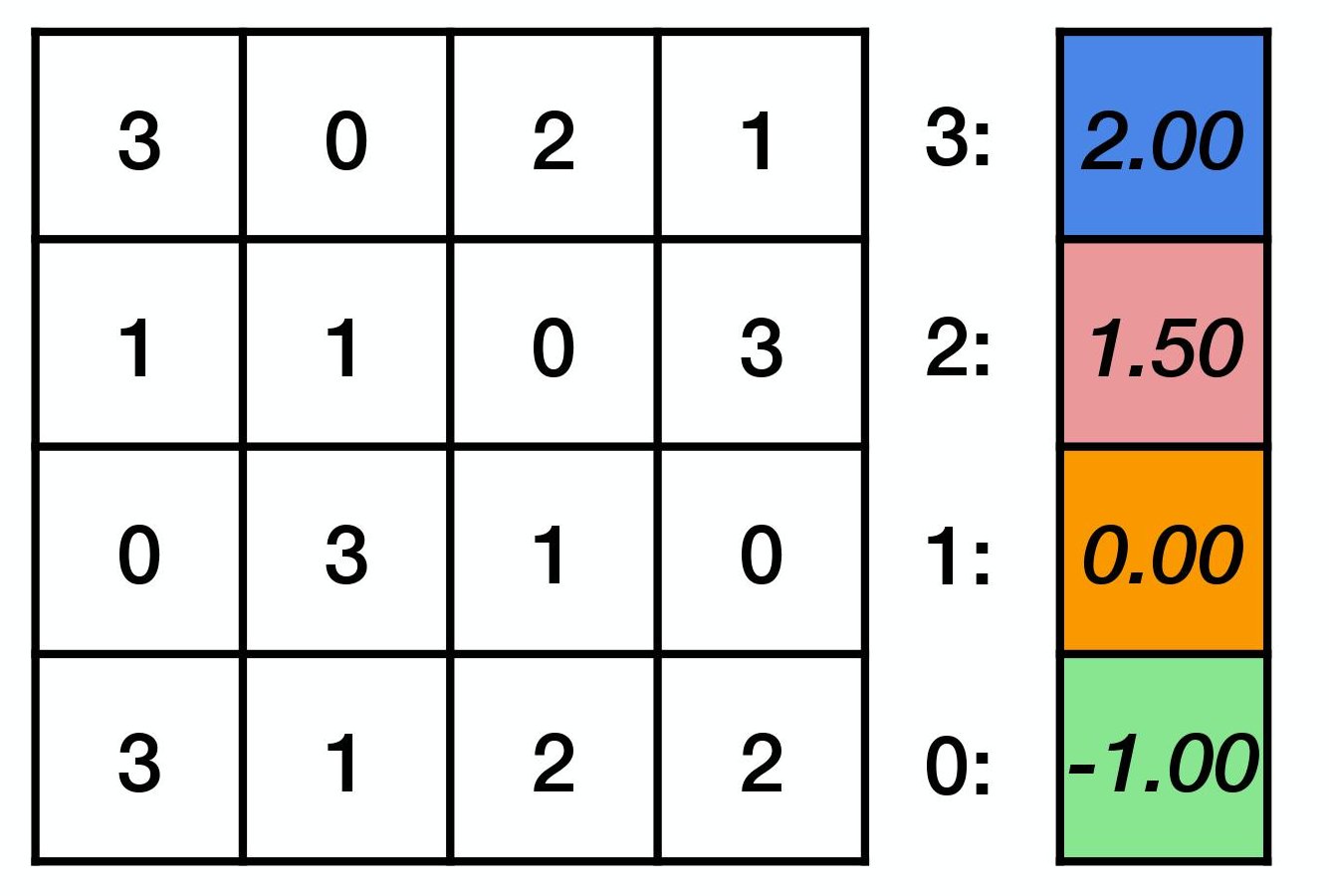}}\quad
    \subfloat[]{\includegraphics[width=0.2\linewidth, height=0.128\linewidth]{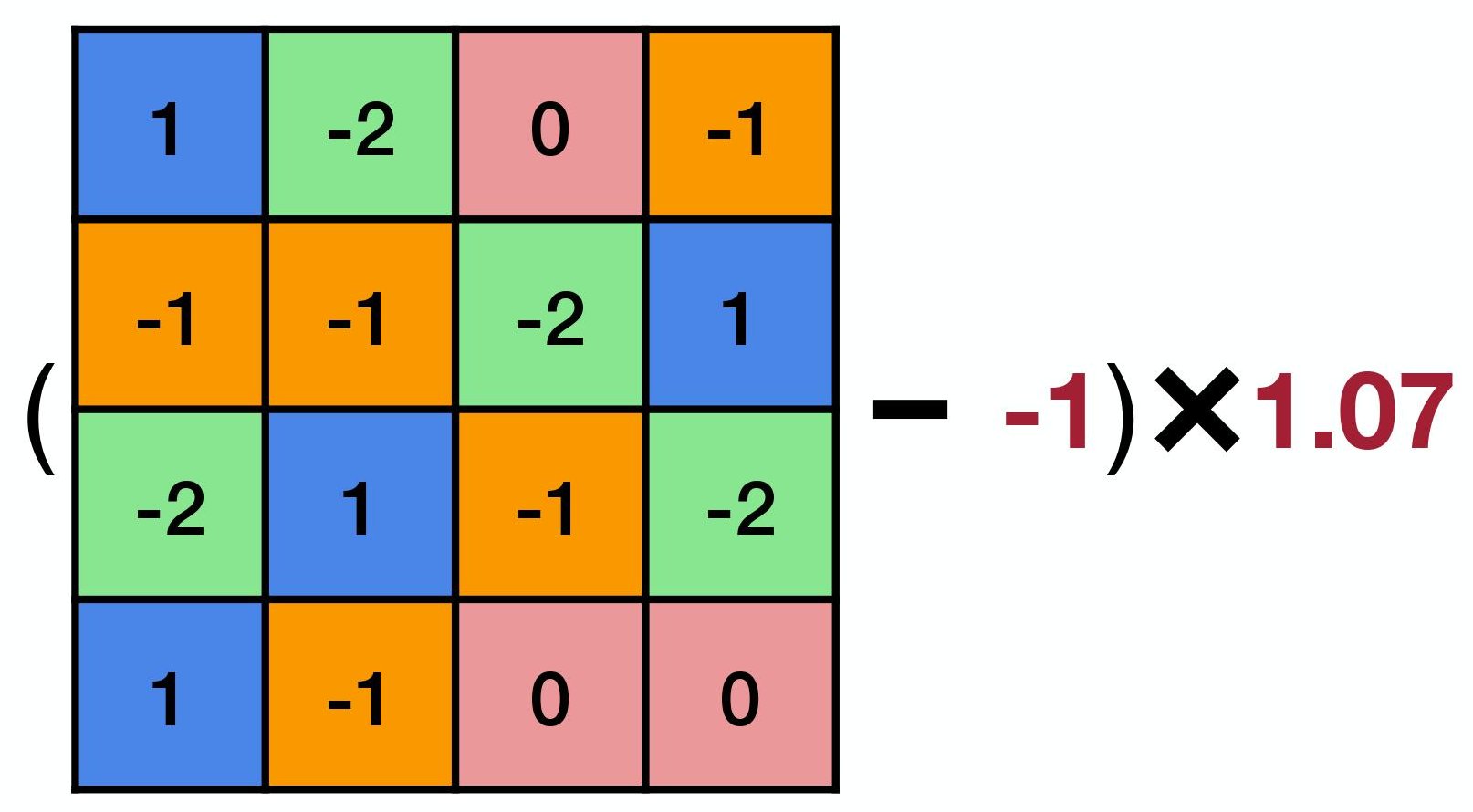}}\\
    
    \subfloat[]{\includegraphics[width=0.323\linewidth, height=0.1\linewidth]{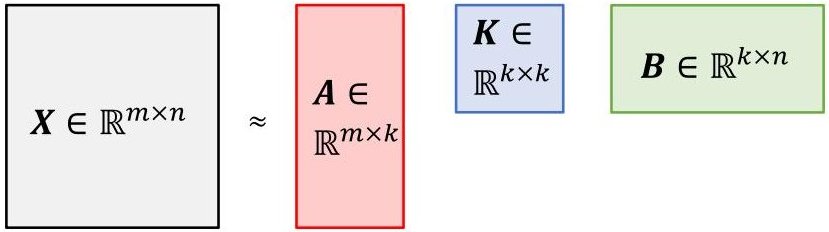}}\qquad
    \subfloat[]{\includegraphics[width=0.323\linewidth, height=0.1\linewidth]{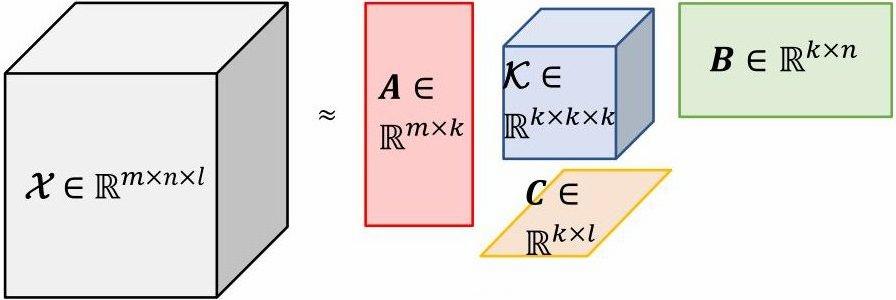}}
\captionsetup{width=0.8\linewidth}
\caption{Illustration of the commonly used model compression techniques.
(a)--(d) Some widely used handcrafted convolutional blocks: (a) Bottleneck convolution \cite{he2015DeepRL}, (b) Grouped convolution \cite{zhang2018shufflenet}, (c) Inverted bottleneck convolution \cite{sandler2018mobilenetv2}, and (d) Depthwise separable convolution \cite{chollet2017} (\textbf{FM} is feature map). (e)--(f) Some popular automatically learned cells: (e) NASNet \cite{zoph2018learning} and (f) Auto-DeepLab \cite{liu2019auto}. (g)--(h) Pruning and sparsification techniques: (g) Weight pruning and (h) Neuron pruning. (i) Knowledge distillation technique \cite{bucica2006}. (j)--(m) Quantisation and binarisation techniques:  (j) Weights matrix, (k) Binary quantisation, (l) K-means-based quantisation, and (m) Linear quantisation. (n)--(o) Low-rank approximation techniques: (n) Low-rank matrix decomposition with $k \times k$ kernel and (o) Low-rank tensor decomposition with $k \times k \times k$ kernel. (Figures are adapted from \cite{deng2020model, zoph2018learning, liu2019auto, deng2020model, menghani2023, efficientDNNslectures, deng2020model}).}
\label{fig:fig2-7}
\end{figure*}

\textbf{Compact architecture:} The fundamental approach to optimize a model for efficient inference is to develop it with building blocks that allow for greater efficiency in computation and memory footprints \cite{menghani2023}. Convolutional layers are largely used building blocks \cite{ma2018ECCV,sandler2018mobilenetv2,chen2019allyou,radosavovic2020designing}, thus, convolution operations contribute most of the computation and memory footprints of a model \cite{wu2019efficient}. Reducing the number of FLOPs in the convolution operations of a model can significantly improve its efficiency \cite{wu2019efficient}. Several handcrafted convolutional building blocks have been proposed to reduce the number of FLOPs in convolution operations, and are widely used to develop models for efficient inference \cite{deng2020model,mi2022survey}. They include depthwise separable convolution \cite{chollet2017}, bottleneck convolution \cite{he2015DeepRL}, inverted bottleneck convolution \cite{sandler2018mobilenetv2}, dilated convolution \cite{fisher2016}, grouped convolution \cite{alex2012,zhang2018shufflenet}, and asymmetric convolution \cite{szegedy2016rethink}. The handcrafted designs have achieved remarkable success, however, handcrafted engineering is laborious and extremely depends on human expertise \cite{zoph2017neural}. Hence, recent years have seen a growing trend for automated machine learning (AutoML) and neural architecture search (NAS) \cite{elsken2019neural,ren2021survey} for automatically searching for efficient cells (i.e., directed acyclic graph of convolutional layers) to develop efficient models \cite{liu2022survey,mi2022survey}. Fig.~\ref{fig:fig2-7}(a)--(d) illustrates some examples of handcrafted convolutional blocks and Fig.~\ref{fig:fig2-7}(e)--(f) shows some automated-learned cells commonly used to develop efficient deep neural networks to reduce the models' number of parameters and FLOPs.

\textbf{Pruning \& sparsification:} These techniques are used to remove ``unimportant" weights and neurons in a model to reduce its number of computation arithmetic operations and make it smaller \cite{luo2017thinet,pavlo2017,mishra2021,huang2019} (see Fig.~\ref{fig:fig2-7}(g)--(h)). This results in a reduction of memory access, thus, accelerating inference. Pruning algorithms have been applied at a different granularity of sparsity, from fine-grained (unstructured) to coarse-grained (structured) \cite{mao2012gran}. For fine-grained pruning, the individual weights or neurons of a model are removed \cite{zhang2018systematic,hu2016NetworkTA,yu2018nisp}, whereas coarse-grained pruning removes entire filters or channels \cite{wen2016,he2017channel}. Unlike fine-grained, the coarse-grained pruning methods are flexible to implement efficiently on hardware but can degrade the model quality \cite{han2017efficient}, however, the pruned model can be retrained to compensate the quality lost \cite{liu2017learning,han2015learning}. AutoML has also been employed to automatically prune the weights and neurons of a model \cite{he2018amc}.

\textbf{Knowledge distillation:} The main idea is to compress a large model by transferring the knowledge of the large model, often called the teacher model, into a small model, also known as the student model \cite{bucica2006,hinton2015dist} (see Fig.~\ref{fig:fig2-7}(i)). Basically, training a small model using a large model as a supervisor to enable the small model, which has less memory and energy footprints, to mimic the quality of the large model for efficient inference \cite{ba2014deep,you2017learn,kim2018ft}. The learning algorithms of knowledge distillation techniques in the literature \cite{wang2022dk,meng2022mdk} can be grouped as self-distillation \cite{furlanello2018}, online distillation \cite{zhang2018dml}, and offline distillation \cite{romero2014fitnets}. Recently, NAS has been adopted in knowledge distillation \cite{li2020kd-nas,kang2019towards}. We refer readers to \cite{gou221kd} for a detailed discussion on this subject.

\textbf{Quantisation \& binarisation:} While pruning reduces the number of weights and neurons in a model, quantisation methods aim to reduce the bit-width of the values of weights and neurons \cite{deng2020model} (see Fig.~\ref{fig:fig2-7}(j)--(m)). This normally renders a model with low-bit precision operands \cite{hubara2017quantized,mishra2018wrpn}. Quantisation methods do not reduce the number of computation arithmetic operations as in pruning but simplify the operations, which therefore can shrink the model size to reduce memory and energy footprints, and to speed up inference but with some sacrifice of model quality \cite{krishnamoorthi2018}. The commonly used methods in the literature \cite{liang2021survey,qin2020bnn,gholami2021survey} include linear quantisation methods \cite{jacob2018quantization}, K-means methods \cite{song2016deepcomp}, and binary/ternary methods \cite{zhu2017trained,yue2021}. Besides these conventional methods, automated quantisation methods (i.e., quantisation with AutoML) have been proposed as well \cite{wang2019haq,lou2019AutoQB}. It is not easy to implement quantisation in practice because it requires a good understanding of bitwise algorithm and hardware architecture \cite{sze2020efficient}. 

\textbf{Low-rank approximation:} Here, the core idea is to reduce the computational complexity of a model by approximating the redundant weight matrices or tensors of a convolutional or fully connected layer using a linear combination of fewer weights \cite{roberto2013,tai2016lowrank}. Compressing a model in this fashion results in a significant reduction in the model size, the model footprint, and the inference latency \cite{jaderberg2014speeding} (see Fig.~\ref{fig:fig2-7}(n)--(o)). Various methods adopted in the literature \cite{junkyu2021,yu2018mc,sze2020efficient} include singular value decomposition (SVD) \cite{jian2014svd}, and tensor decomposition methods such as Tucker decomposition \cite{young2016comp,song2020speed} and canonical polyadic (CP) decomposition \cite{lebedev2015}.

\subsection{Efficiency Metrics}\label{sec:2.1}
Other than the quality of segmentation results, an efficient model has three core characteristics: \textit{smaller}, \textit{faster}, and \textit{greener}, which are respectively gauged on \textit{storage}, \textit{inference speed}, and \textit{energy} \cite{menghani2023,schwartz2020green}. When designing or choosing a deep neural network architecture for semantic segmentation in real-time applications, one major consideration is the quality-cost trade-off \cite{paleyes2022}. Accuracy is used to measure the quality of model results. The most commonly used accuracy evaluation metric in semantic segmentation is the intersection over union (IoU), known as Jaccard index, and its variants mean IoU (mIoU) and frequency-weighted IoU (FwIoU) \cite{takos2020,ulku2022} (see Equations~\ref{eq:iou}, \ref{eq:miou}, and \ref{eq:fwiou}). The model footprint (i.e., computational cost), typically associated with the training and deploying of a deep learning model is a very important factor in considering a method for real-time semantic segmentation \cite{menghani2023}. Most often, the more computational budget a model is given, the more quality its segmentation results and vice versa \cite{dehghani2022}. 

\begin{equation}\label{eq:iou}
    IoU = \frac{\sum_{j=1}^{k} n_{jj}}{\sum_{j=1}^{k} (n_{ij} + n_{ji} + n_{jj})}, \qquad i\neq j
\end{equation}
\begin{equation}\label{eq:miou}
    mIoU = \frac{1}{k} \sum_{j=1}^{k} \frac{n_{jj}}{(n_{ij} + n_{ji} + n_{jj})}, \qquad i\neq j
\end{equation}
\begin{equation}\label{eq:fwiou}
    FwIoU = \frac{1}{\sum_{j=1}^{k}t_j} \sum_{j=1}^{k} {t_j} \frac{n_{jj}}{(n_{ij} + n_{ji} + n_{jj})}, \quad i\neq j
\end{equation}
where $n_{jj}$ is the number of pixels labelled as class $j$ and classified as class $j$, $n_{ij}$ is the number of pixels labelled as class $i$ but classified as class $j$ (false positive), and $n_{ji}$ is the number of pixels labelled as class $j$ but classified as class $i$ (false negative). The FwIoU is an improved version of mIoU which weighs the importance of each class with $t_j$ depending on its appearance frequency.

To achieve a better quality-cost trade-off, several efficiency metrics have been adopted in the literature of efficient deep learning computing and its applications such as real-time semantic segmentation for measuring model footprint. The efficiency metrics view the computational cost from different complexity perspectives, that is, \textit{memory} and \textit{computation}. The memory-related efficiency metrics include the number of parameters, the model size, and the size of activations, which are normally used to measure storage \cite{bartoldson2022,dehghani2022,menghani2023,dollar2021FastAA,radosavovic2020designing}. The number of parameters metric is the parameter count, that is, the number of elements in the weight tensors of the given model (see Table~\ref{tab:param-cal}). The model size metric is used to measure the storage for the weight tensors of a model. Generally, if the weight tensors of a model have the same data type (e.g., floating point), the model size metric is expressed as:
\begin{equation}
   ModelSize = \#parameters \times bitwidth.
\end{equation}
The common measurement units are kilobytes (KB) and megabytes (MB). For example, a model with 60M parameters stored in 32-bit, has a total storage, i.e., model size, of 240MB (60M $\times$ 4 Bytes). The activations size metric measures the memory requirement for the feature maps or activations of a model during a forward pass (i.e., inference). Like the model size metric, the activations size metric is measured in KB or MB. With a batch size of 1, the size of activations of a model can be expressed as:
\begin{equation}
   ActSize = [\mathit{input\mbox{-}size} + {\sum_{l=1}^{L} (\mathit{\#neurons})^l}]\times{bitwidth},
\end{equation}
where $\#neurons$ is the number of neurons of layer $l$ and $L$ is the number of layers in the model. The $\#neurons$ depends on the type of layer $l$. For example, if $l$ is a convolutional layer, the $\#neurons$ is $c_o\times\ h_o\times w_o,$ where $c_o$ is the output channels, $h_o$ the output height, and $w_o$ is the output width. And for a linear layer, $\#neurons$ is simply the output channels $c_o$.

\begin{table}
\centering
\captionsetup{width=0.98\linewidth}
\caption{The computation of the number of parameters and multiply-accumulate (MAC) operations of a neural network layer. The $c_i$ is input channels, $c_o$ is output channels, $h_o$ is output height, $w_o$ is output width, $k_h$ is kernel height, $k_w$ is kernel width, and $g$ is split groups.}
\label{tab:param-cal}
\scalebox{0.93}{
\begin{tabular}{l c c}
    \hline\hline
    \multirow{2}{*}{Layer} & \#Parameters & MACs \\
     & (bias is ignored) & (batch size $n=1$) \\
    \hline
    Linear & $c_o \cdot c_i$ & $c_o \cdot c_i$ \\
    Convolution & $c_o \cdot c_i \cdot k_h \cdot k_w$ & $c_o \cdot c_i \cdot k_h \cdot k_w \cdot h_o \cdot w_o$ \\
    Grouped Convolution  & $c_o \cdot c_i \cdot k_h \cdot k_w/g$ & $c_o \cdot c_i \cdot k_h \cdot k_w \cdot h_o \cdot w_o/g$ \\
    Depthwise Convolution & $c_o \cdot k_h \cdot k_w$  & $c_o \cdot k_h \cdot k_w \cdot h_o \cdot w_o$ \\
    \hline\hline
\end{tabular}}
\vspace{-2mm}
\end{table}

Computation-related metrics are the number of multiply-accumulate (MAC) operations and the number of floating-point operations (FLOPs) \cite{dehghani2022,menghani2023}. An operation that computes the product of two numbers and adds the result to an accumulator is considered one MAC operation. For example, the MACs of a matrix-vector product $Ax$ is expressed as:
\begin{equation}
    MACs = m \cdot n, \quad A \in \mathcal{R}^{m\times n}, \quad x \in\mathcal{R}^n.
\end{equation}
For matrix-matrix product $AB$, the MACs is expressed as:
\begin{equation}
    MACs = m \cdot n \cdot p, \quad A \in \mathcal{R}^{m\times n}, \quad B \in\mathcal{R}^{n\times p}.
\end{equation}
See Table~\ref{tab:param-cal} for the MACs count of various neural network layers. In FLOPs, a multiply is one operation and an addition is also one operation, hence, one MAC operation is two FLOPs. For example, a model with 724M MACs has 1.4G FLOPs (724M $\times$ 2). Furthermore, FLOPs per second (FLOPS) can be expressions:
\begin{equation}
    FLOPS = \frac{FLOPs}{second}.
\end{equation}

\begin{figure}[!t]
\centering
\includegraphics[width=0.9\linewidth, height=0.45\linewidth]{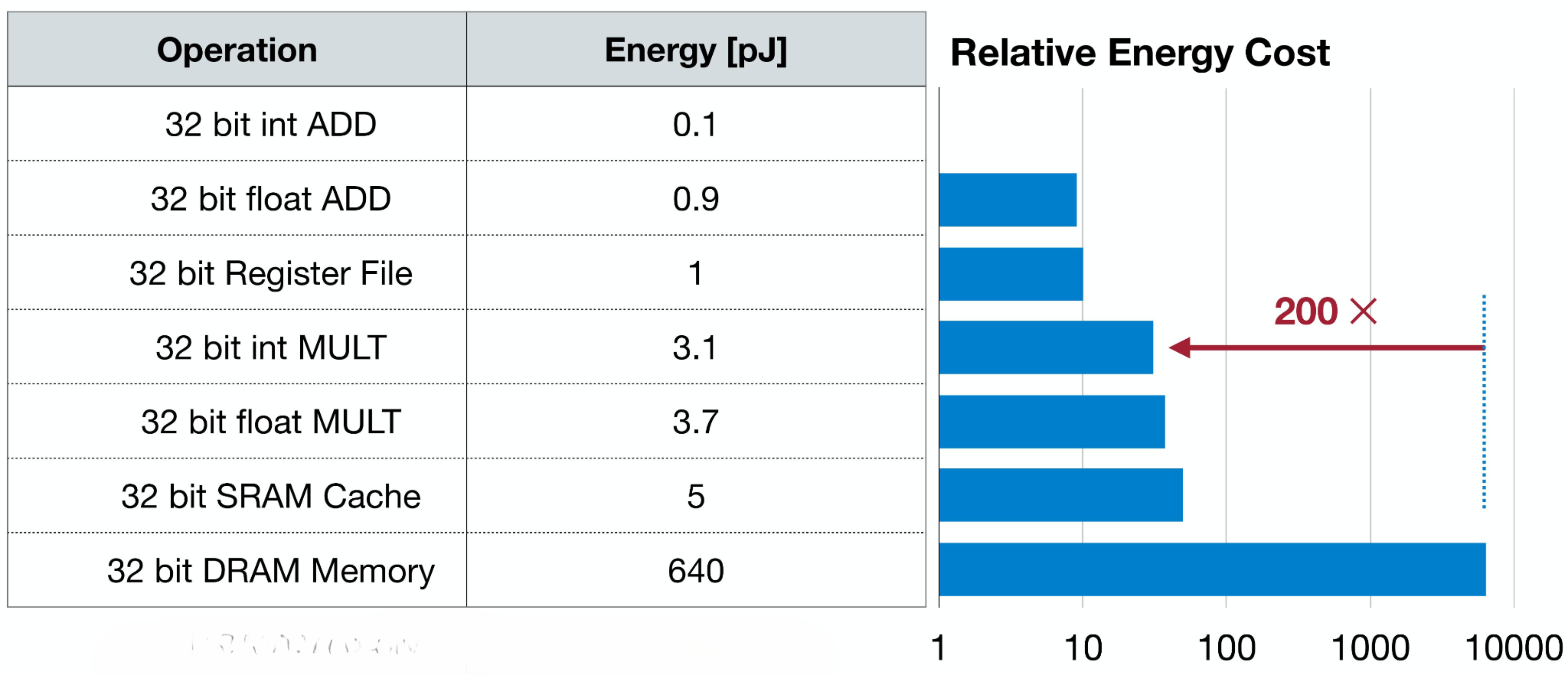}
\captionsetup{width=0.93\linewidth}
\caption{The energy cost for various arithmetic and memory access operations in a 45 nanometer (nm) process. Data movement consumes significantly higher energy than arithmetic operations. (Figure adapted from \cite{efficientDNNslectures}).}
\label{fig:energy}
\end{figure}

\begin{figure}[!t]
\vspace{2mm}
\centering
\includegraphics[width=0.78\linewidth, height=0.45\linewidth]{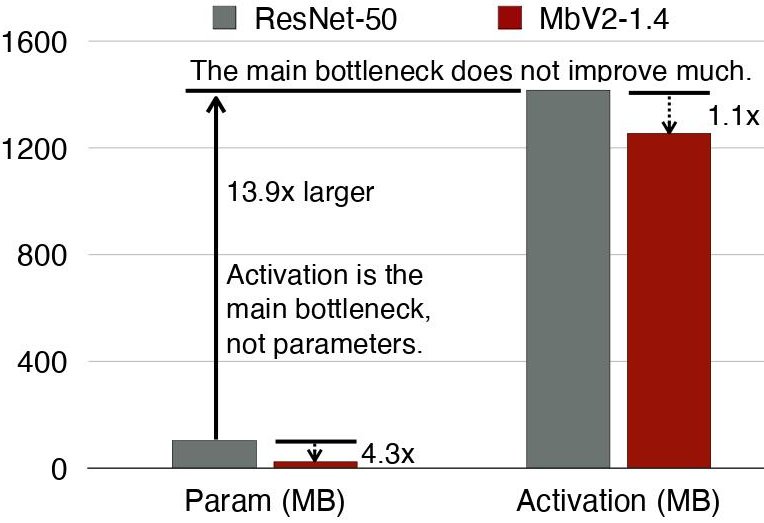}
\captionsetup{width=0.9\linewidth}
\caption{The memory cost comparison under batch size 16 of ResNet-50 (larger model) and MobileNetV2-1.4 (compact model). A reduction in the number of parameters (model size) does not reduce the activation size, which is the main memory bottleneck. (Figure adapted from \cite{cai2020}).}
\label{fig:memory}
\vspace{-2mm}
\end{figure}

\begin{table*}[!t]
\centering
\captionsetup{width=0.85\linewidth}
\caption{List of efficient deep neural networks adopted in real-time semantic segmentation of remote sensing imagery. The networks are categorized into handcrafted designs and automated neural architecture search (NAS) of efficient deep neural networks in remote sensing.}
\label{tab:methods_rs}
\begin{tabular}{c c c c c}
    \hline\hline
    Year & Method & Datasets & Application & Reference \\
    \hline 
    \multicolumn{5}{c}{\textit{Handcrafted models}} \\
    \hline
    2019& SegRBM-Net & UAV & Ground vehicle segmentation&\cite{KHOSHBORESHMASOULEH2019172} \\
    2020 & AlexNet & UAV & Weed mapping & \cite{app10207132} \\
    2021 & MobileNetV2-UNet/FFB-BiSeNetV2 & UAV & Weed monitoring & \cite{rs13214370} \\
    2021 & Context aggregation network & UAV & Semantic segmentation & \cite{YANG2021124}  \\
    2021& ABCNet & Aerial images & Semantic segmentation &\cite{LI202184} \\
    2022 & BASNet & UAV & Burned area segmentation & \cite{9852471} \\
    2022 & LWCDnet & Landsat/MODIS & Cloud detection & \cite{9775756} \\
    2022 & RSR-Net &  Aerial images & Building segmentation & \cite{9627995} \\
    2022 & MSL-Net & Aerial images & Building segmentation & \cite{qiu2022} \\
    2022 & LightFGCNet & Aerial images & Building/Semantic segmentation & \cite{rs14246193} \\
    2022& CF-Net & Aerial images & Semantic segmentation & \cite{9340601} \\
    2022& SGBNet & Google earth images & Semantic segmentation &\cite{doi:10.1080/01431161.2021.2022805} \\
    2022 & U-NetFormer & Aerial images & Semantic segmentation&  \cite{WANG2022196}\\
    2022& LPASS-Net &Aerial images & Semantic segmentation & \cite{rs14236057} \\
    2022& DSANet & Aerial images & Semantic segmentation& \cite{rs14215399} \\
    2022& SegFormer &Aerial images & Semantic segmentation & \cite{rs14051294} \\
    2023 & LRAD-Net & Aerial images & Building segmentation & \cite{jiabin2023} \\
    \hline
    \multicolumn{5}{c}{\textit{NAS models}} \\
    \hline
    2020& NAS-HRIS & Gaofen multispectral/Aerial& Building/Semantic segmentation & \cite{rs20185292} \\
    2021& SFPN & 3D point clouds & Semantic segmentation & \cite{LIN2021279} \\
    2022& DNAS & Gaofen multispectral& Semantic segmentation & \cite{DNAS2022} \\
    2022& U-Net-like search space & Worldview-2 & Semantic segmentation & \cite{MauricioISPRS2022} \\
    2022 & Evolutionary NAS & Aerial images & Semantic segmentation & \cite{DBLP:journals/nca/Broni-BediakoMM22} \\
    \hline\hline
\end{tabular}
\end{table*}

The inference speed and energy indicators are difficult to estimate compared to the storage indicators because they depend on both the model's architecture and the hardware platform the model is deployed on \cite{sze2020efficient}. In many instances, the inference speed of a model is reported as throughput \cite{sze2020efficient} and latency \cite{wu2019fbnet}. Throughput measures how much data can be processed or how many executions of a task can be completed in a given amount of time (i.e., images processed per second), whereas latency indicates the period of time between when input data arrives and when the result is generated by a model. Achieving high throughput and low latency is critical for real-time semantic segmentation applications. However, in some cases, it may be difficult to achieve high throughput and low latency simultaneously, e.g., increasing throughput via batching of multiple images will also increase the latency of a model \cite{sze2020efficient}. From queuing theory \cite{john1961}, the relationship between throughput and latency can be expressed as:
\begin{equation}
    Throughput = \frac{\#tasks}{Latency},
\end{equation}
where $\#tasks$ is the number of input images in a batch and $Latency$ is the latency of the model which is determined by:
\begin{equation}
    Latency = max(\mathcal{T}_{computation}, \mathcal{T}_{memory}),
\end{equation}
where $\mathcal{T}_{memory}$ is the number of operations in a model divided by the number of operations that a machine (e.g., GPU or CPU processor) can process per second, and $\mathcal{T}_{memory}$ is the total data movement time of activations and weights into memory, which depends on the memory bandwidth of the processor. In Ma \textit{et al.}~\cite{ma2018ECCV}, the number of memory accesses is reported as a surrogate measure for model inference speed. On the other hand, power consumption \cite{you2022zeus} and carbon emission footprint \cite{schwartz2020green} have been used to measure the energy efficiency of a model. Power consumption is used to determine the amount of energy a model consumes per unit of time. More data movement implied more memory access which leads to high energy consumption \cite{horowitz20141} (see Fig.~\ref{fig:energy}). Real-time applications in a resource-constrained environment with limited power capacity require a model with low energy consumption (i.e., high energy efficiency). For inference, energy efficiency is typically reported as inferences per joule \cite{sze2020efficient}.  

In the literature, there is a common misconception that the computational cost (efficiency) indicators are correlated. For example, fewer model parameters mean less memory footprint. Cai~\textit{et al.} \cite{cai2020} demonstrated that the number of activations is the memory bottleneck during training or during inference, but not the number of parameters (see Fig.~\ref{fig:memory}). Also, a reduction in FLOPs does not necessarily translate into a reduction in latency~\cite{jierun2023}. Thus, to devise or adopt a compact model for real-time semantic segmentation, it is proper to evaluate the efficiency using several computational cost indicators because there is no single efficiency metric that is sufficient \cite{dehghani2022}. We refer to \cite{bartoldson2022,dehghani2022,dollar2021FastAA} for a detailed discussion on this topic.

\section{Real-Time Segmentation Models}\label{sec:3}
This section summarizes the state-of-the-art efficient deep neural networks for real-time semantic segmentation of remote sensing imagery (see Table~\ref{tab:methods_rs}). Most of the real-time semantic segmentation models adapt one of the widely used compact backbone networks, including MobileNet~\cite{sandler2018mobilenetv2}, SqueezeNet~\cite{iandola2017squeezenet}, ShuffleNet~\cite{zhang2018shufflenet}, and  EfficientNet~\cite{tan2019efficientnet} which were designed for image classification tasks. Large-scale architectures such as ResNet~\cite{he2015DeepRL}, U-Net~\cite{ronneberger2015u}, VGG~\cite{Simonyan15}, and ViT~\cite{dosovitskiy2021an} have also been compressed and adapted for real-time semantic segmentation. The segmentation tasks include building extraction \cite{9627995}, burned detection \cite{9852471}, weed mapping \cite{rs13214370}, cloud detection \cite{9775756}, and others \cite{YANG2021124,9340601}. And most of the works were developed for unmanned aerial vehicle (UAV) platforms \cite{app10207132, KHOSHBORESHMASOULEH2019172}. Here, the presentation is grouped into models built with handcrafted architectures and the ones that were developed via automated NAS (AutoML). 

\subsection{Handcrafted Models}\label{sec:3.1}
Deng \textit{et al.}~\cite{app10207132} introduced a lightweight weed mapping via real-time image processing onboard a UAV based on AlexNet~\cite{alex2012}. The authors established a hardware environment for real-time image processing that integrates map visualization, flight control, and image collection onboard a UAV for time-efficient weed mapping. The BiSeNetV2 \cite{fan2021rethinking} and U-Net were employed as backbone networks in Lan \textit{et al.}~\cite{rs13214370} to build two identification models, MobileNetV2-UNet and FFB-BiSeNetV2, for real-time analysis of UAV low-altitude imagery for timely monitoring of rice weeds in farmland. BASNet~\cite{9852471} proposed an efficient burned area segmentation network with ResNet as a backbone network to improve the performance of UAV high-resolution image segmentation. In LWCDnet~\cite{9775756}, a backbone of a ResNet-like style module was used to build a lightweight auto-encoder model for cloud detection using Landsat 8 and MODIS datasets. Huang \textit{et al.}~\cite{9627995} introduced RSR-Net, a lightweight network based on U-Net and SqueezeNet, to extract buildings from remote sensing images of WHU building dataset \cite{ji2018buildings}. In MSL-Net~\cite{qiu2022}, MobileNet architectural module (depthwise separable convolution) with atrous spatial pyramid pooling (ASPP)~\cite{chen2017deeplab} as a multiscale feature extractor was used to alleviate network performance degradation of building extraction. LRAD-Net~\cite{jiabin2023} employed depthwise separable convolutions with ASPP and self-attention mechanism \cite{Vaswani2017} to achieve state-of-the-art performance on WHU building datasets. Yang \textit{et al.}~\cite{YANG2021124} introduced a context aggregation network, a dual-branch convolutional neural network based on MobileNet, which has the potential to capture both global aggregation and local contextual dependencies that are required for accurate semantic segmentation using low computational overheads. Chen \textit{et al.}~\cite{rs14246193} proposed a lightweight global context semantic segmentation network called LightFGCNet. The authors employed a U-Net-like framework to fully utilize the global context data and adapted the spatial pyramid pooling (SPP)~\cite{he2014spatial} as a strategy to reduce the number of network parameters. CF-Net~\cite{9340601} adopted VGG and ResNet to develop a cross-fusion network for fast and effective extraction of multiscale semantic information, especially for small-scale semantic information. RT-SegRBM-Net~\cite{KHOSHBORESHMASOULEH2019172}, for semantic segmentation of ground vehicles from UAV-based thermal infrared imagery, the performance of deep learning combined model of SegNet \cite{badrinarayanan2017segnet} was improved using the Gaussian-Bernoulli restricted Boltzmann machine \cite{kyung2013GBRBM}. In Liu \textit{et al.}~\cite{siyu2021}, EfficientNet was used as a backbone to build a lightweight model with fewer parameters for semantic segmentation of UAV remote sensing imagery. For real-time segmentation of land cover, SGBNet~\cite{doi:10.1080/01431161.2021.2022805} employed a semantics-guided strategy with a bottleneck network to balance accuracy and inference speed. 

Besides the aforementioned conventional convolutional neural networks, generative adversarial networks (GAN)~\cite{GoodfellowGAN2014} and Transformers~\cite{dosovitskiy2021an} have been adapted for real-time semantic segmentation of remote sensing imagery. In~\cite{land10010079}, the challenge of post-processing semantic segmentation predictions of road surface area was tackled using a conditional GAN based on pix2pix \cite{isola2017image} to obtain state-of-the-art performance. Wang \textit{et al.}~\cite{WANG2022196}, introduced a Transformer-based network constructed in a U-Net-like fashion (U-NetFormer) for segmenting urban scenes in real time. ABCNet~\cite{LI202184} adopted the attention mechanism and followed the design concept of BiSeNet to build a lightweight model that retains rich spatial details and captures global contextual information. In~\cite{rs14236057}, the authors proposed an end-to-end lightweight progressive attention semantic segmentation network (LPASS-Net) based on MobileNet with an attentional feature fusion network. LPASS-Net aims to solve the problem of computational cost reduction without sacrificing segmentation accuracy. DSANet~\cite{rs14215399} introduced an effective deep supervision-based attention network with spatial and enhancement loss functions for real-time semantic segmentation. Yan \textit{et al.}~\cite{rs14051294} adapted SegFormer~\cite{xie2021segformer} to develop an efficient depth fusion transformer network, which downsamples input with patch merging strategy and utilizes a depth-aware self-attention module for effective aerial image segmentation.

\subsection{Neural Architecture Search Models}\label{sec:3.2}
The commonly used search strategies in the automated NAS literature~\cite{elsken2019neural} include reinforcement learning \cite{zoph2017neural}, differentiable search (gradient-based algorithm) \cite{liu2018darts}, evolutionary algorithms \cite{liu2021survey}, and Bayesian optimisation \cite{White2019BANANASBO}. Most of the proposed NAS methods in remote sensing adopted the differentiable search strategy because it is faster by a large order of magnitude and uses less computational search time compared to reinforcement learning or evolutionary algorithms ~\cite{heuillet2023efficient}. In~\cite{DNAS2022}, a hierarchical search space of three levels (path-level, connection-level, and cell-level) was designed to build a supernet space. A differentiable search strategy was designed to automatically search for lightweight networks for high-resolution remote sensing image semantic segmentation. de Paulo \textit{et al.}~\cite{MauricioISPRS2022} used the algorithm in \cite{DNAS2022} to improve the design of a U-Net-like network search space. The authors replaced 3x3 convolution layers with parallel layers of different kernel sizes. Then they pruned the network with a scaled sigmoid strategy \cite{guo2022differentiable} for multispectral satellite image semantic segmentation. SFPN~\cite{LIN2021279} also used a differentiable search strategy on a discrete search space to generate a feature pyramid supernet to search a feature pyramid network module for 3D point cloud semantic segmentation. NAS-HRIS~\cite{rs20185292} embeds a novel directed acyclic graph into the DARTS~\cite{liu2018darts} search space to learn end-to-end cell modules, normal and reduction cells, using gradient descent optimization. The searched cells are used to design a semantic segmentation network for high-resolution satellite images. In~\cite{DBLP:journals/nca/Broni-BediakoMM22}, an evolutionary NAS method was introduced in the semantic segmentation of high-resolution aerial images. The authors leverage the complementary strengths of gene expression programming and cellular encoding to build a search space to evolve a modularized encoder-decoder network via an evolutionary process.

\begin{table}[t]
\captionsetup{width=0.85\linewidth}
\caption{The number and proportion of pixels and the number of segments of the eight classes~\cite{Xia2023WACV}.}
\label{tab:oem_pixels_segments}
\begin{center}
\scalebox{1}{\begin{tabular}{c c c c c}
    \hline\hline
    Colour & \multirow{2}{*}{Class} & \multicolumn{2}{c}{Pixels}& Segments\\
    \cline{3-4} (HEX) & & Count (M) & (\%) & (K) \\
    \hline
    \colorbox{bareland}{\textcolor{white}{800000}} & Bareland  & 74   & 1.5 & 6.3 \\
    \colorbox{rangeland}{\textcolor{white}{00FF24}}& Rangeland & 1130 & 22.9 & 459.4 \\
    \colorbox{develop}{\textcolor{white}{949494}}& Developed space  & 798  & 16.1 & 382.7 \\
                                FFFFFF & Road   & 331  & 6.7 & 27.9 \\
    \colorbox{tree}{\textcolor{white}{226126}} & Tree & 996  & 20.2 & 902.9\\
    \colorbox{water}{\textcolor{white}{0045FF}}& Water & 161  & 3.3 & 18.7 \\
    \colorbox{agriculture}{\textcolor{white}{4BB549}}& Agriculture land & 680  & 13.7 & 18.2 \\
    \colorbox{building}{\textcolor{white}{DE1F07}}& Building & 770  & 15.6 & 389.3 \\
    \hline\hline
    \end{tabular}}
    \vspace{-4mm}
\end{center}
\end{table}

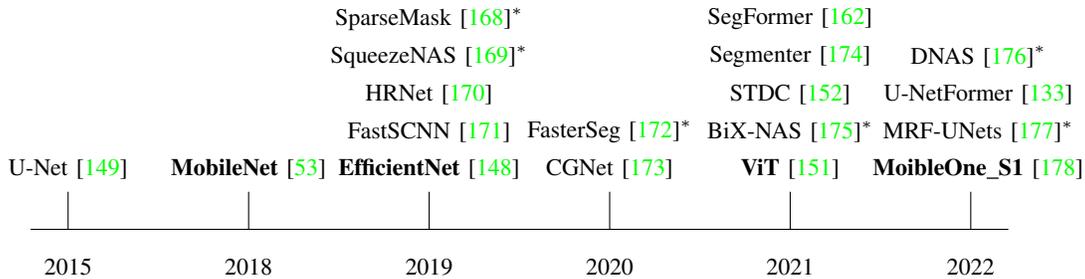
\begin{figure*}[t]
\centering
\small
\begin{tikzpicture}
    \draw (0,0)--(13,0);
    \draw (0.5,0)--(0.5,0.5);
    \draw (2.9,0)--(2.9,0.5);
    \draw (5.3,0)--(5.3,0.5);
    \draw (7.7,0)--(7.7,0.5);
    \draw (10.1,0)--(10.1,0.5);
    \draw (12.5,0)--(12.5,0.5);
    \node at (0.5,0.8) {U-Net~\cite{ronneberger2015u}};
    \node at (0.5,-0.5) {2015};
    \node at (2.9,0.8) {\textbf{MobileNet}~\cite{sandler2018mobilenetv2}};
    \node at (2.9,-0.5) {2018};
    \node at (5.3,2.8) {SparseMask~\cite{Wu2019SparseMask}$^\ast$}; 
    \node at (5.3,2.3) {SqueezeNAS~\cite{2019_SqueezeNAS}$^\ast$}; 
    \node at (5.3,1.8) {HRNet~\cite{SunXLW19}}; 
    \node at (5.3,1.3) {FastSCNN~\cite{poudel2019fast}};
    \node at (5.3,0.8) {\textbf{EfficientNet}~\cite{tan2019efficientnet}};
    \node at (5.3,-0.5) {2019};
    \node at (7.7,1.3) {FasterSeg~\cite{Chen2020FasterSeg}$^\ast$};
    \node at (7.7,0.8) {CGNet~\cite{wu2020cgnet}};
    \node at (7.7,-0.5) {2020};
    \node at (10.1,2.8) {SegFormer~\cite{xie2021segformer}};
    \node at (10.1,2.3) {Segmenter~\cite{strudel2021segmenter}};
    \node at (10.1,1.8) {STDC~\cite{fan2021rethinking}};
    \node at (10.1,1.3) {BiX-NAS~\cite{wang2021bix}$^\ast$};
    \node at (10.1,0.8) {\textbf{ViT}~\cite{dosovitskiy2021an}};
    \node at (10.1,-0.5) {2021};
    \node at (12.6,2.3) {DNAS~\cite{rs14163864}$^\ast$};
    \node at (12.6,1.8) {U-NetFormer~\cite{WANG2022196}};
    \node at (12.6,1.3) {MRF-UNets~\cite{Wang2022MRF-UNets}$^\ast$};
    \node at (12.6,0.8) {\textbf{MoibleOne\_S1}~\cite{mobileone2022}};
    \node at (12.5,-0.5) {2022};
\end{tikzpicture}
\captionsetup{width=0.8\linewidth}
\caption{The list of efficient deep neural networks we adopted for the comparative study on the OpenEarthMap benchmark. The \textit{$\ast$} indicates automated architecture search networks and the boldface denotes the backbone networks of the handcrafted methods.}
\vspace{-2mm}
\label{fig:list-of-methods}
\end{figure*}

\section{Comparative Study on OpenEarthMap}\label{sec:4}
\subsection{OpenEarthMap Description}\label{sec:4.1}
The OpenEarthMap~\cite{Xia2023WACV} is a sub-meter dataset for high-resolution land cover mapping at a worldwide scale. The dataset is made up of 5000 aerial and satellite images at a ground sample distance of 0.25-0.5m with hand-annotated 8-class land cover labels of 2.2 million segments. It covers 97 areas that are spread out across 44 countries on 6 continents. OpenEarthMap shows the diversity and intricacy of satellite and UAV segmentation. Table~\ref{tab:oem_pixels_segments} displays the quantity and ratio of pixels labelled for each class and the number of segments identified for each category. It also reveals semantic segmentation feasibility issues under challenging scenarios. Following the experimental settings in \cite{Xia2023WACV}, we evaluate representative efficient DNNs models on semantic segmentation and continent-wise domain generalisation tasks. The 5000 images were randomly divided into training, validation, and test sets with a ratio of 6:1:3 for each region, yielding 3000, 500, and 1500 images for the semantic segmentation task. For the continent-wise domain generalisation, the data for each continent is treated as a source domain, while the other continents are considered target domains. 

\subsection{Compared Methods}\label{sec:4.2}
This section details the efficient neural networks, as shown in Fig.~\ref{fig:list-of-methods}, which we evaluate on the OpenEarthMap as part of this work. We consider both handcrafted and automated architecture-searched methods for real-time semantic segmentation. The following are the handcrafted networks we employ for the study: U-Net series, U-NetFormer series, FastSCNN, SegFormer, Segmenter, HRNet, STDC, and CGNet. On the side of automated architecture-searched methods, the following six networks are used: SqueezeNAS,  BiX-NAS, MRF-UNets, DNAS, SparseMask, and FasterSeg. A brief introduction of the handcrafted networks is as follows:

\textbf{U-Net}~\cite{ronneberger2015u} series: We develop U-Net baseline architectures identical to the original U-Net architecture. It consists of lightweight encoder blocks for downsampling, decoder blocks for upsampling, concatenation blocks that fuse low- and high-level features, and skip connections for improving accuracy. The lightweight encoder blocks also reduce computation cost and improve inference speed. Our main contribution is to benchmark the accuracy and efficiency of U-Net using various lightweight encoders. We design asymmetric U-Net baselines with lightweight encoders of MobileNet~\cite{sandler2018mobilenetv2}, MoibleOne\_S1~\cite{mobileone2022}, and EfficientNet-B0 and B1 ~\cite{tan2019efficientnet}.

\begin{figure}[h]
\centering
\includegraphics[width=0.75\linewidth, height=0.73\linewidth]{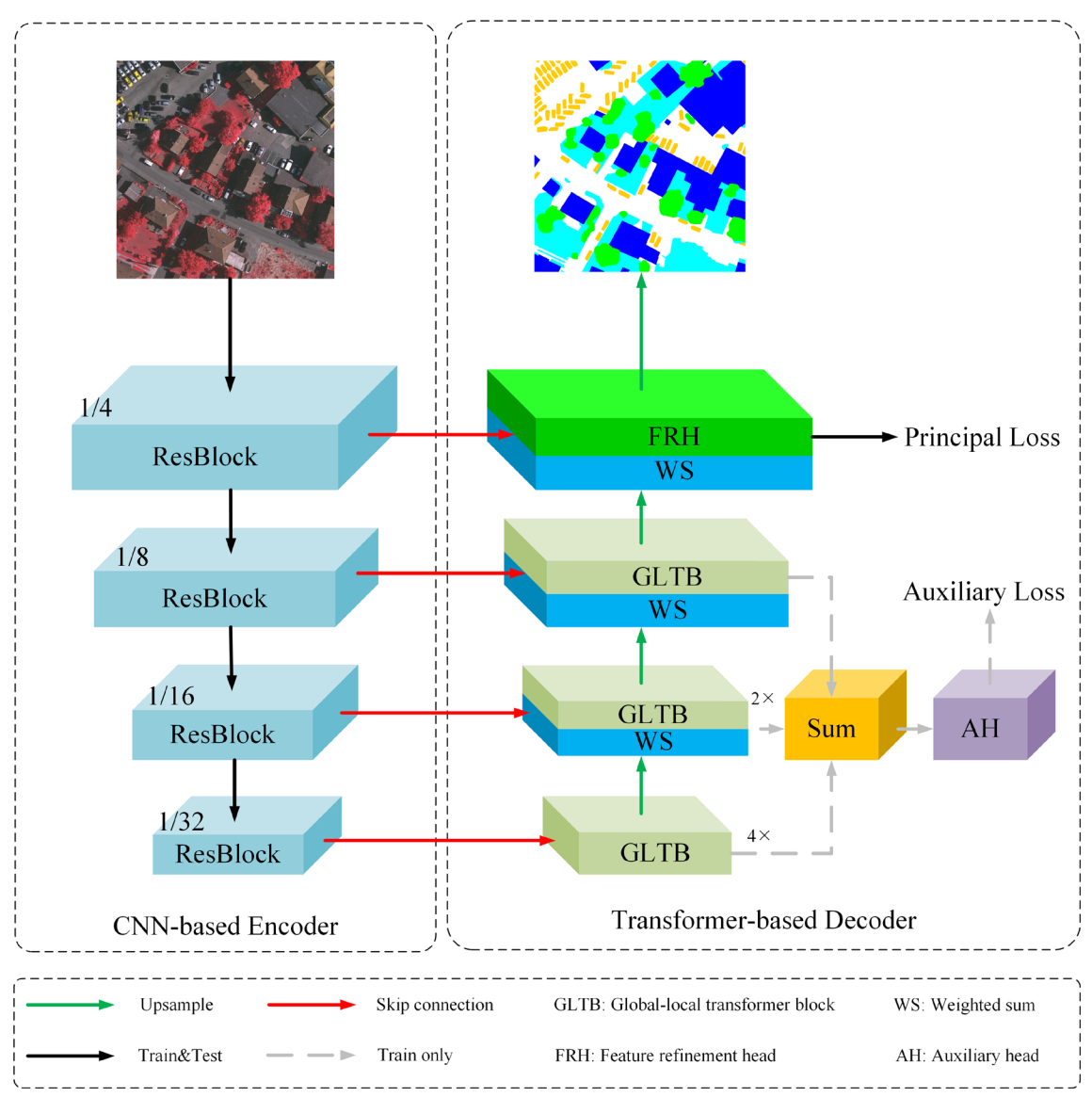}
\captionsetup{width=0.85\linewidth}
\caption{The architecture overview the main idea of U-NetFormer. (Figure adapted from \cite{WANG2022196}).}
\label{fig:unetformer}
\vspace{-1mm}
\end{figure}

\textbf{U-NetFormer}~\cite{WANG2022196} series: The main novelty here is the structure of the decoder. The decoder is a transformer-based architecture that efficiently models global and local information. The decoder’s global–local transformer block (GLTB) constructs two parallel branches to extract the global and local contexts (see Fig~\ref{fig:unetformer}). The semantic features generated by the encoder are fused with the features from GLTB to produce generalized fusion features. In short, a transformer-based decoder captures global and local contexts at multiple scales and maintains high efficiency. For U-NetFormer, we also considered EfficientNet versions B0, B1, and B2 ~\cite{tan2019efficientnet}.

\begin{figure}[!h]
\centering
\includegraphics[width=0.95\linewidth, height=0.35\linewidth]{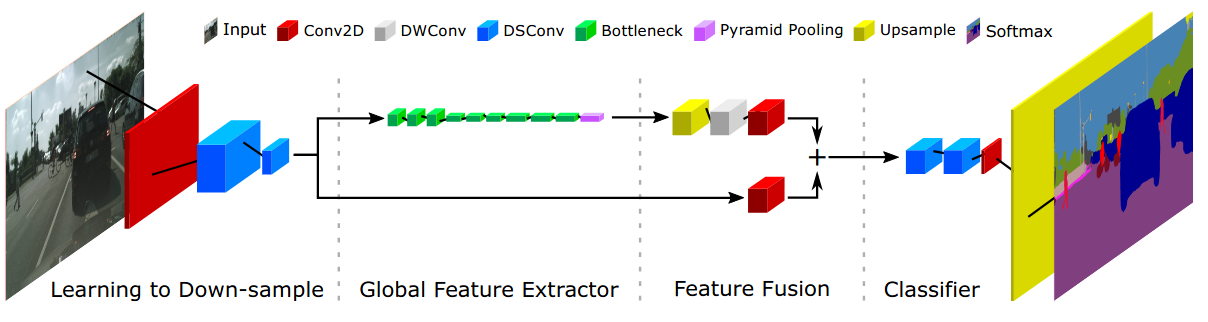}
\captionsetup{width=0.95\linewidth}
\caption{An illustration of the main idea of Fast-SCNN. It shares the computations between branches of encoders to build a segmentation network. (Figure adapted from \cite{poudel2019fast}).}
\label{fig:fastscnn}
\vspace{-2mm}
\end{figure}

\textbf{FastSCNN}~\cite{poudel2019fast}: The FastSCNN is suitable for efficient computation on embedded devices with low memory. Building on existing two-branch methods for fast segmentation,  the "learning to downsample" module, which computes low-level features for multiple resolution branches simultaneously is the main idea of FastSCNN (see Fig.~\ref{fig:fastscnn}). It combines high-resolution spatial details with deep features extracted at a lower resolution.

\begin{figure}[!h]
\centering
\includegraphics[width=0.95\linewidth, height=0.45\linewidth]{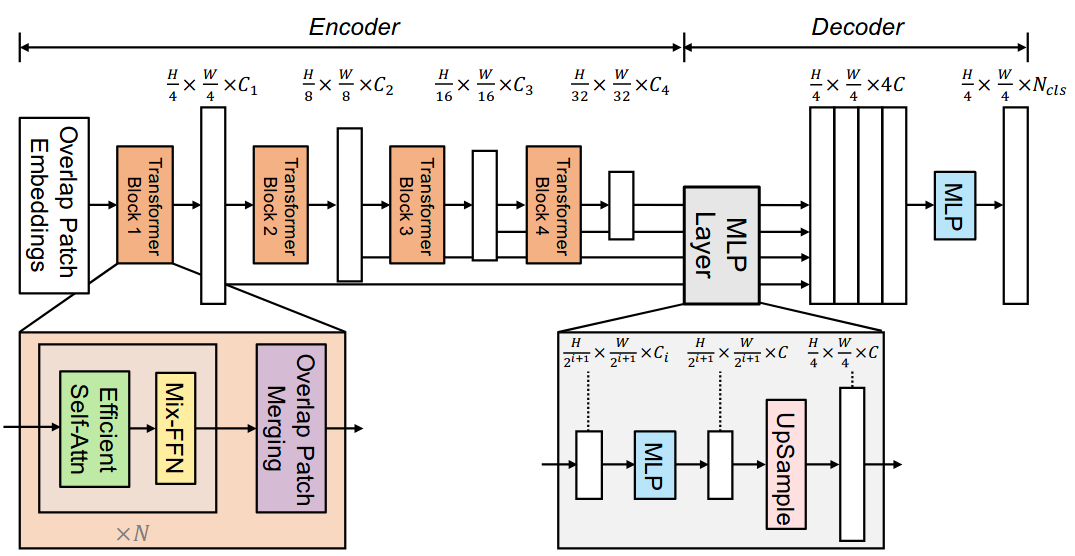}
\captionsetup{width=0.95\linewidth}
\caption{The SegFormer framework consists of two modules: A hierarchical Transformer encoder which extracts coarse and fine features, and a lightweight MLP decoder which fuses multi-level features and predicts the semantic segmentation mask. The “FFN” indicates a feed-forward network (Figure adapted from \cite{xie2021segformer}).}
\label{fig:segformer}
\end{figure}

\textbf{SegFormer}~\cite{xie2021segformer}: The encoder comprises multiple mix transformer (MiT) encoders, i.e., MiT-B0–MiT-B5, allowing the encoder to obtain multilevel features to generate the segmentation mask. Furthermore, this structure enables the MLP decoder to combine local and global attention to generate effective representations. The hierarchical transformer encoder has a larger effective receptive field than conventional CNN encoders, allowing lightweight MLP as a decoder (see  Fig.~\ref{fig:segformer}). This model enables the scalability of the network such that we change the transformer layers according to our resources. In this work, we avoid using large networks such as MiT-B1–MiT-B5, instead, we use MiT-B0 to improve efficiency in real-time applications.

\begin{figure}[!h]
\centering
\includegraphics[width=0.9\linewidth, height=0.4\linewidth]{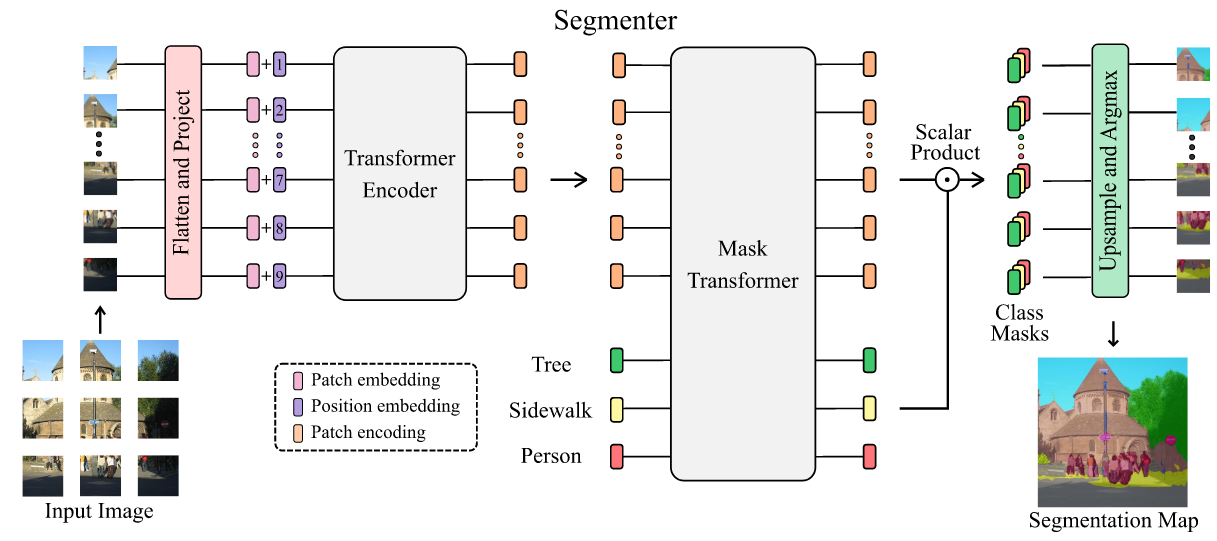}
\captionsetup{width=0.95\linewidth}
\caption{An overview of the Segmenter framework. Encoder (left): image patches are projected to an embedding sequence, and then encoded with a transformer. Decoder (right): a mask transformer receives the output of the encoder and class embeddings as input to predict a segmentation mask. (Figure adapted from \cite{strudel2021segmenter}).}
\label{fig:segmenter}
\vspace{1mm}
\end{figure}

\textbf{Segmenter}~\cite{strudel2021segmenter}: The Segmenter network relies on the output embeddings corresponding to image patches and obtains class labels from these embeddings with a point-wise linear decoder or a mask Transformer decoder (see Fig~\ref{fig:segmenter}). This work considers the tiny Vision Transformer (ViT)~\cite{dosovitskiy2021an}.

\begin{figure}[!h]
\centering
\includegraphics[width=0.95\linewidth, height=0.25\linewidth]{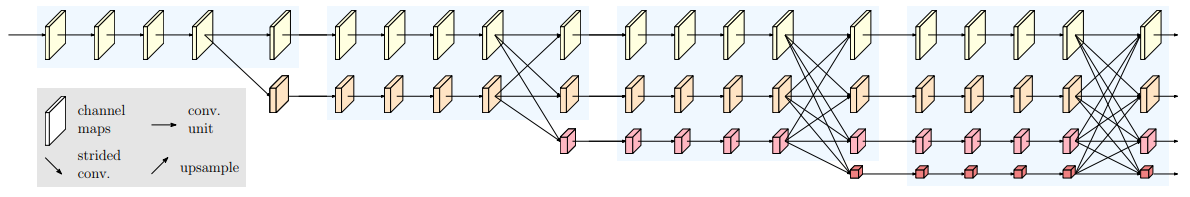}
\captionsetup{width=0.95\linewidth}
\caption{Illustration of the main idea of the HRNet architecture. Consists of high-to-low resolution parallel subnetworks with multi-scale fusion. The horizontal direction indicates the depth of the network, and the vertical indicates the scale of the feature maps. (Figure adapted from \cite{SunXLW19}).}
\label{fig:hrnet}
\vspace{1mm}
\end{figure}

\textbf{HRNet}~\cite{SunXLW19}: The high-resolution net (HRNet) can maintain high-resolution representations throughout the whole process. HRNet starts from a high-resolution convolution stream, gradually adds high-to-low-resolution convolution streams one by one, and connects the multi-resolution streams in parallel (see Fig.~\ref{fig:hrnet}). The resulting network consists of several stages (4 in the original paper), and the $n$th stage contains streams corresponding to resolutions. In this work, we use the HRNet-small version.

\begin{figure}[!h]
\vspace{-1mm}
\centering
\includegraphics[width=0.95\linewidth, height=0.45\linewidth]{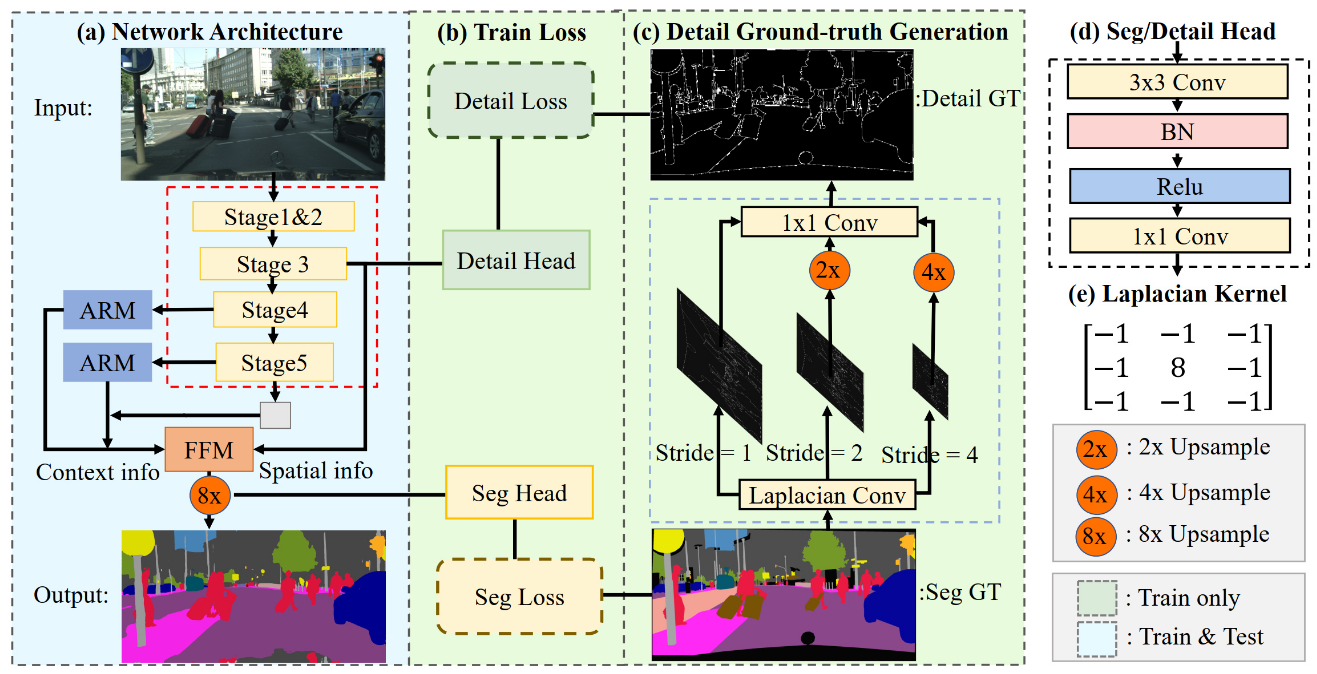}
\captionsetup{width=0.95\linewidth}
\caption{The architecture overview of the STDC network. ARM is attention refine module and FFM is feature fusion module. The dashed blue box indicates the detail aggregation module. The dashed red box is the network proposed in STDC. (Figure adapted from \cite{fan2021rethinking}).}
\label{fig:stdc}
\end{figure}

\textbf{STDC}~\cite{fan2021rethinking}: The short-term dense concatenate network (STDC) is a novel and efficient structure for removing structure redundancy. Specifically, the network gradually reduces the dimension of feature maps and uses their aggregation for image representation, forming the STDC network's basic module. In the decoder, a detail aggregation module is proposed by integrating the learning of spatial information into low-level layers in a single-stream manner. Finally, the low-level features and deep features are fused to predict the final segmentation results (see Fig.~\ref{fig:stdc}).

\begin{figure}[!h]
\centering
\includegraphics[width=0.95\linewidth, height=0.35\linewidth]{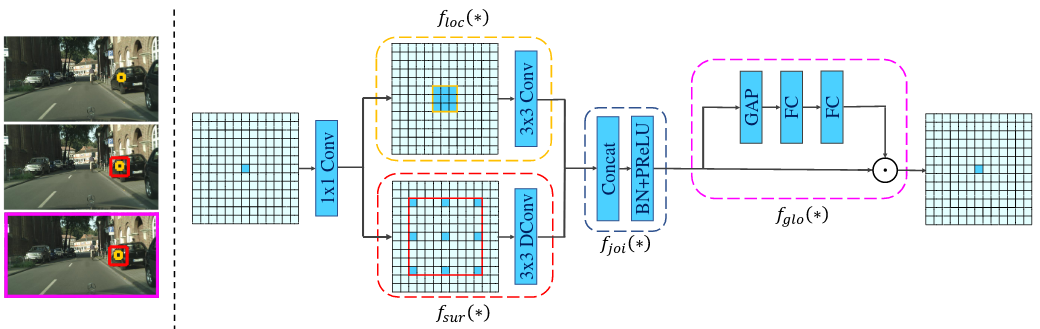}
\captionsetup{width=0.95\linewidth}
\caption{The architecture overview of the context-guided block in CGNet. The structure consists of a local feature extractor \textit{floc($\ast$)}, a surrounding context extractor \textit{fsur($\ast$)}, a joint feature extractor \textit{fjoi($\ast$)}, and a global context extractor \textit{fglo($\ast$)}. (Figure adapted from \cite{wu2020cgnet}).}
\label{fig:cgnet}
\vspace{-2mm}
\end{figure}

\textbf{CGNet}~\cite{wu2020cgnet}: It is a lightweight network for semantic segmentation.  The CGNet uses a context-guided block to learn joint features of both local and surrounding contexts (see Fig.~\ref{fig:cgnet}). Based on the context-guided block, CGNet can capture contextual information in all stages of the network, specially tailored for increasing segmentation accuracy. CGNet is also elaborately designed to reduce the number of parameters and save memory footprint.

We also present a brief introduction of the NAS methods that are employed for the study:\par
\begin{figure}[!h]
\vspace{-1mm}
\centering
\includegraphics[width=0.95\linewidth, height=0.3\linewidth]{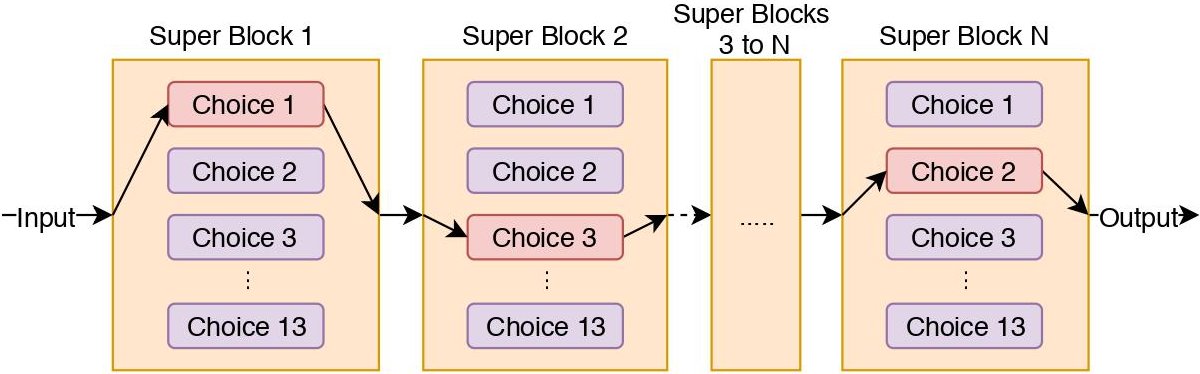}
\captionsetup{width=0.95\linewidth}
\caption{An overview of an architecture path of a sampled architecture from a supernetwork. In the 1st superblock, candidate block 1 is selected, then the 2nd superblock selects candidate block 3, and the \textit{N}th superblock selects candidate block 2. (Figure adapted from \cite{2019_SqueezeNAS}).}
\label{fig:squeezenas}
\end{figure}

\textbf{SqueezeNAS}~\cite{2019_SqueezeNAS}: The SqueezeNAS is considered the first proxyless hardware-aware search which was targeted for dense semantic segmentation. Using a differentiable search strategy, it advances the state-of-the-art accuracy for latency-optimized networks on the Cityscapes~\cite{Cordts2015Cvprw} semantic segmentation dataset via a search space similar to MobileNet. An overview of the SqueezeNAS architecture search path is shown in Fig.~\ref{fig:squeezenas}.

\begin{figure}[!h]
\centering
\includegraphics[width=0.95\linewidth, height=0.35\linewidth]{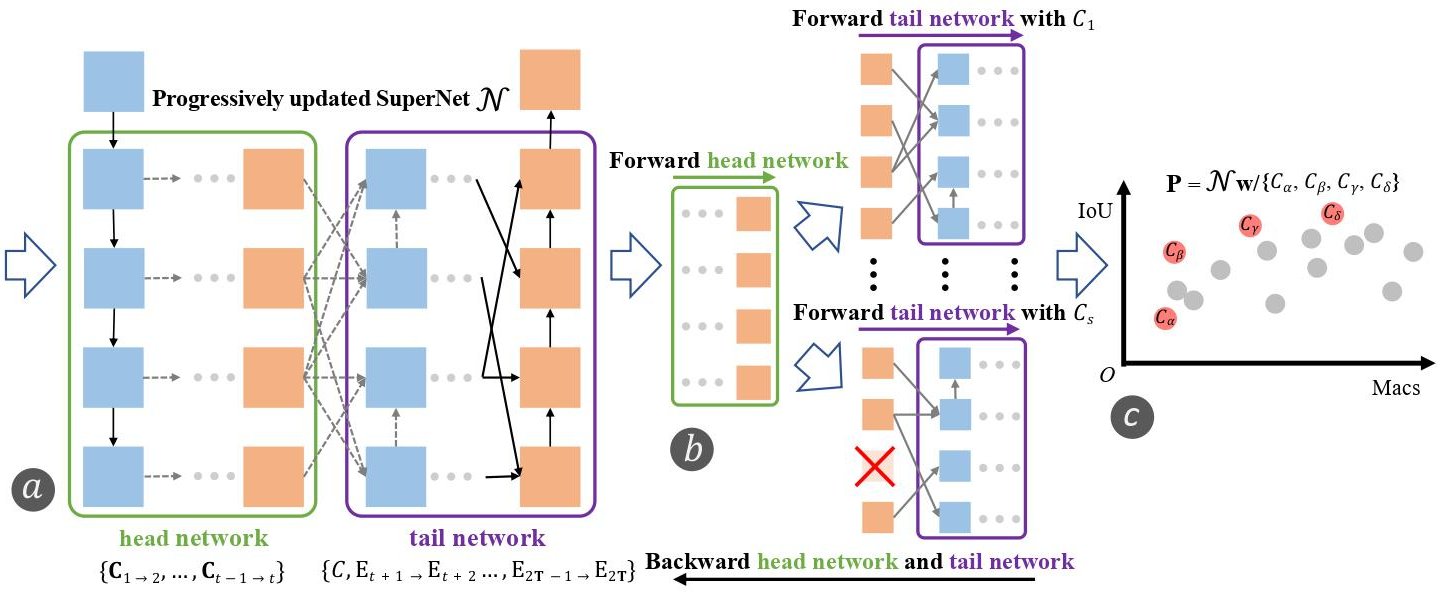}
\captionsetup{width=0.95\linewidth}
\caption{The BiX-NAS progressive evolutionary search overview. (a) Phase 1: searched supernet $\mathcal{N}$ divided into a head and a tail networks. (b) Proposed forward and backward schemes. (c) Phase 2: searched skips at the Pareto front of \textbf{P} are only retained. (Figure adapted from \cite{wang2021bix}).}
\label{fig:bixnas}
\end{figure}

\textbf{BiX-NAS}~\cite{wang2021bix}: The BiX-NAS is based on a multi-scale upgrade of a bi-directional skip-connected network (see Fig.~\ref{fig:bixnas}). It uses a two-phase search algorithm, a differentiable search in Phase1, and an evolutionary search in Phase2. It reduces network computational costs by sifting out ineffective multi-scale features at different levels and iterations.

\begin{figure}
\centering
\includegraphics[width=0.95\linewidth, height=0.55\linewidth]{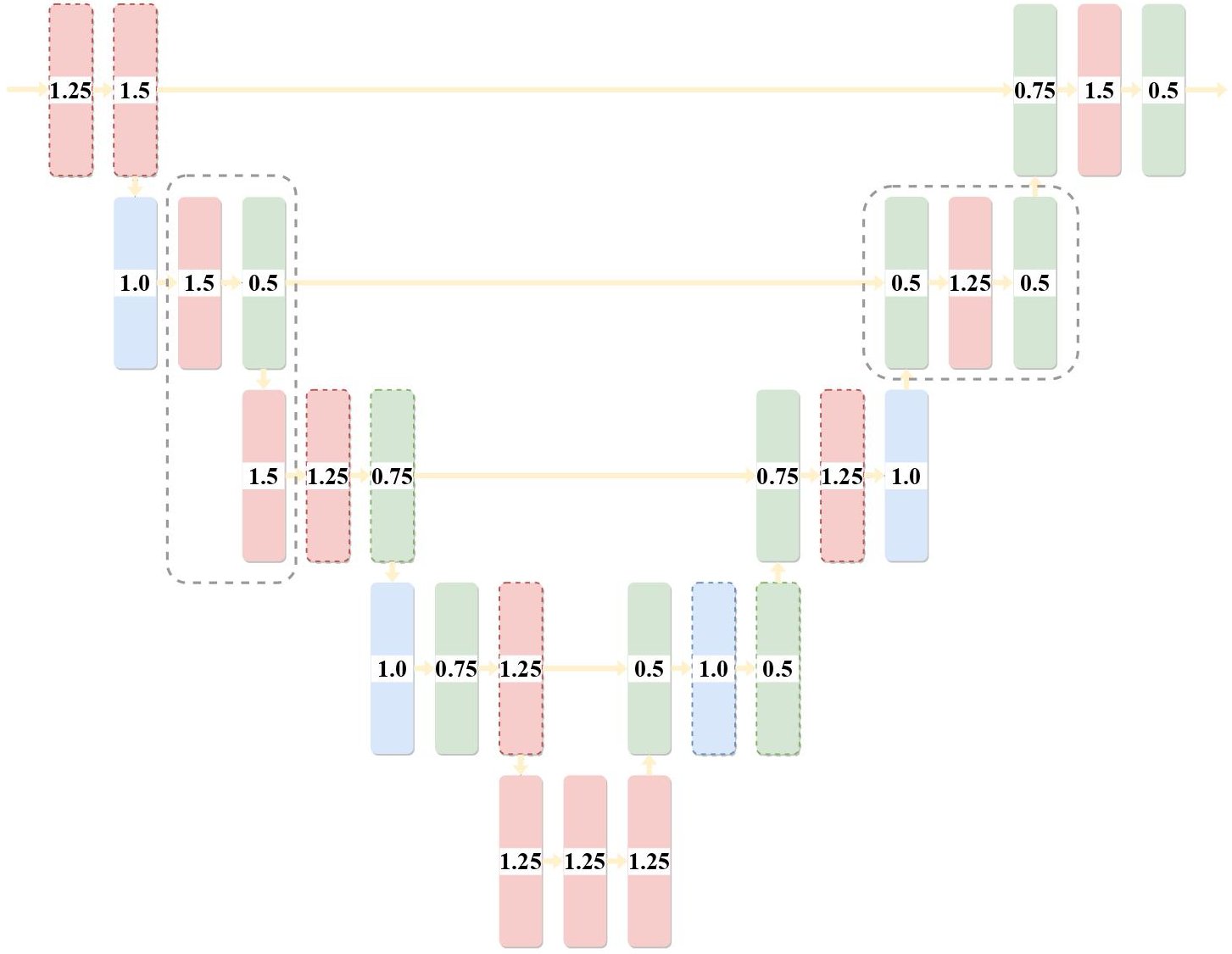}
\captionsetup{width=0.95\linewidth}
\caption{The architecture structure of MRF-UNets. The width ratios assigned to the original U-Net are the values inside rectangles. Rectangles with coloured dashed lines use 5$\times$5
kernels and the others use 3$\times$3 kernels. On the left, the grey dashed line represents an example of a bottleneck block in the encoder. On the right, the grey dashed line shows an example of an inverted bottleneck block in the decoder. (Figure adapted from \cite{Wang2022MRF-UNets}).}
\label{fig:mrfunet}
\vspace{-2mm}
\end{figure}

\textbf{MRF-UNets}~\cite{Wang2022MRF-UNets}: It extends and improves the recent adaptive and optimal network width search (AOWS)~\cite{berman2020aows} method with a more general Markov random field (MRF) framework, a diverse M-best loopy inference \cite{batra2012}, and a differentiable parameter learning. This provides the necessary NAS framework to efficiently explore the architecture search space of a U-Net backbone (see Fig.~\ref{fig:mrfunet}).

\begin{figure}[!h]
\centering
\includegraphics[width=0.9\linewidth, height=0.5\linewidth]{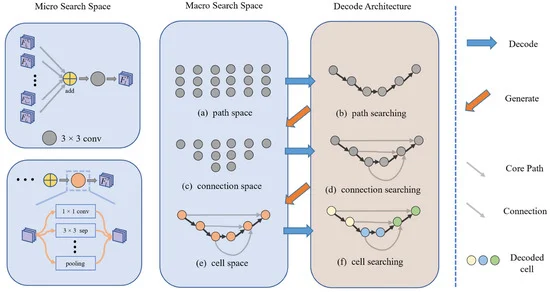}
\captionsetup{width=0.95\linewidth}
\caption{ The decoupling neural architecture search framework. It consists of micro search space, macro search space (path-level, connection-level, and cell-level), and decoded architecture. (Figure adapted from \cite{rs14163864}).}
\label{fig:dnas}
\end{figure}

\textbf{DNAS}~\cite{rs14163864}: The decoupling NAS (DNAS) employs a hierarchical search space with three levels: path-level, connection-level, and cell-level (see Fig.~\ref{fig:dnas}) to automatically design the network architecture via a differentiable search for HRSI semantic segmentation. In DNAS, the search optimization strategy consists of finding optimal path connections of a super-net for developing a lightweight network.

\begin{figure}
\centering
\includegraphics[width=0.72\linewidth, height=0.45\linewidth]{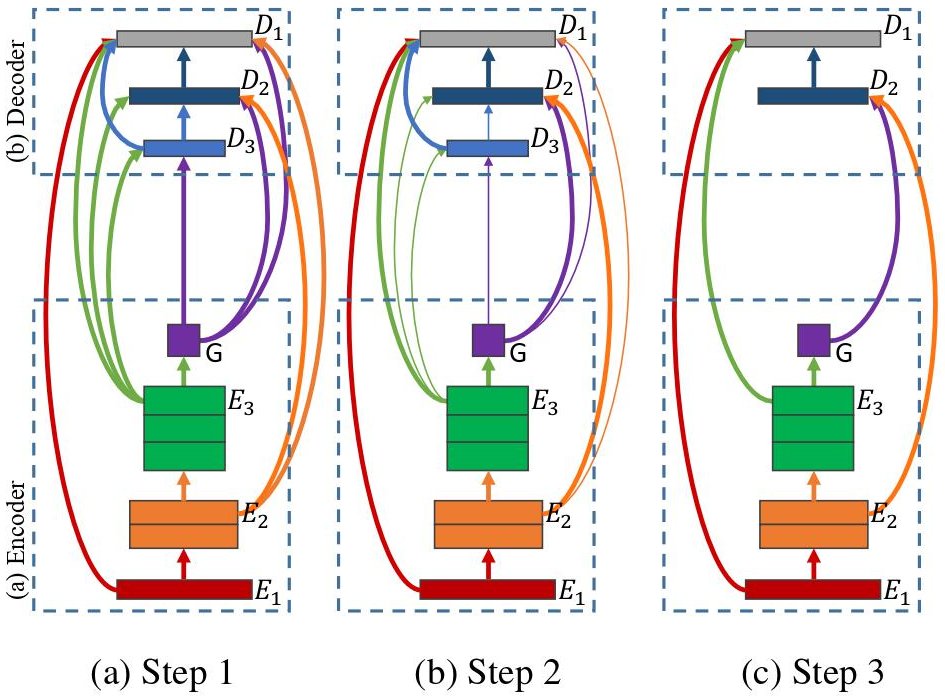}
\captionsetup{width=0.95\linewidth}
\caption{The architecture search framework overview of SparseMask. (a) A pretrained image classifier is transformed into a fully dense network with learnable connections. (b) The optimal connectivity is searched with the proposed sparse loss using a differentiable search strategy. (c) Less important connections are dropped to obtain the final architecture. (Figure adapted from \cite{Wu2019SparseMask}).}
\label{fig:sparsemask}
\vspace{-3mm}
\end{figure}

\textbf{SparseMask}~\cite{Wu2019SparseMask}: The SparseMask creates a densely connected network with learnable connections which contains a large set of possible final connectivity structures as the search space (see Fig.~\ref{fig:sparsemask}). Then, a gradient-based search strategy is employed to search for optimal connectivity from the dense connections by making the connections to be sparse.

\begin{figure}[!h]
\vspace{-2mm}
\centering
\includegraphics[width=0.85\linewidth, height=0.5\linewidth]{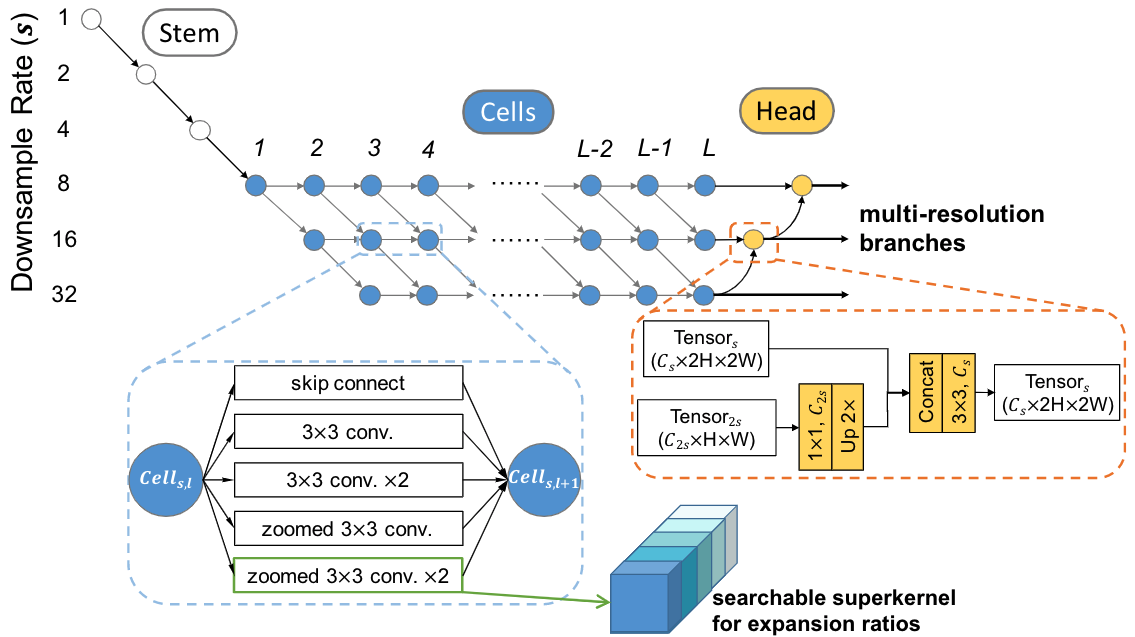}
\captionsetup{width=0.95\linewidth}
\caption{The FasterSeg multi-resolution search space. It aims to optimize multiple branches with different output resolutions. The outputs are progressively aggregated into a head module. Each searchable cell may have two inputs and two outputs with different downsampling rates. For each cell, searching for expansion ratios is enabled within a single super-kernel. (Figure adapted from \cite{Chen2020FasterSeg}).}
\label{fig:fasterseg}
\end{figure}

\textbf{FasterSeg}~\cite{Chen2020FasterSeg}: The FasterSeg is developed from a broader search space integrating multi-resolution branches as shown in Fig.~\ref{fig:fasterseg}, which is commonly used in manually designed segmentation models. To calibrate the balance between the goals of high accuracy and low latency, FasterSeg introduced a decoupled and fine-grained latency regularization to effectively overcome the phenomenons where the searched networks are prone to “collapsing” to low-latency yet poor-accuracy models.

\subsection{Experimental Settings}\label{sec:4.3}
Both the handcrafted and NAS methods we use for the experiments are PyTorch-based. For the handcrafted methods, the U-Net-based architectures are adopted from Yakubovskiy~\cite{Yakubovskiy:2019} and Wang \textit{et al.}~\cite{WANG2022196}, and the other architectures are from MMsegmentation~\cite{mmseg2020}. The networks are trained on a single NVIDIA GPU DGX-1/DGX-2 with 16/32GB of RAM. The number of epochs is set to 200, and a batch size of 8 with an image input size of 512$\times$512 randomly cropped is employed. The cross-entropy (CE) loss is used in training all the networks. For the U-Net-based architectures, we use AdamW optimizer~\cite{loshchilov2018decoupled} with a learning rate of $1\times10^{-4}$ and weight decay of $1\times10^{-6}$. For the MMsegmentation-based architectures, we use the default settings of each method. We adopt stochastic gradient descent (SGD) optimizer with a learning rate of $1\times10^{-3}$, weight decay of $5\times10^{-4}$, and momentum of 0.9 for the HRNet networks. The rest of the networks use AdamW optimizer with a learning rate set as $6\times10^{-5}$, weight decay as 0.01, and betas parameters as 0.9 and 0.999. A polynomial learning rate decay with a factor of 1.0 and an initial linear warm-up of 1500 iterations is used. The backbones in all the handcrafted networks are pre-trained on the ImageNet dataset. No data augmentation is applied during training and testing for all networks. Following previous works \cite{ENet2016,chen2017deeplab,zhao2017pyramid}, we use mIoU to assess the segmentation quality of all the networks. To assess the networks' suitability for real-time applications of remote sensing image semantic segmentation, we also evaluate their inference speed (FPS), i.e., \textit{frames per second}, in addition to computational complexity (FLOPs) and the number of parameters. For the NAS methods, we adopt the code from the GitHub page of each method. With the exception of the evaluation metrics mentioned above, we follow the default experimental settings stated in the papers of each method to search and train the networks. All the NAS experiments are run on the same machines use for the handcrafted ones. 

\subsection{Results and Discussion}\label{sec:5}
The performance evaluation of the efficient deep neural networks on the OpenEarthMap dataset is presented in Table~\ref{tab:segmentation_results}. It provides the class IoU for all eight classes of the OpenEarthMap and the overall segmentation quality in mIoU for each model. Also, it presents the number of trainable parameters (Params), the computational complexity (FLOPs), and the inference speed (FPS) to measure the efficiency of the models. The FLOPs and FPS were computed on 1024x1024 RGB input data. The inference speed was computed on a single NVIDIA Tesla P100 (DGX-1) with 16 GB memory based on an average time of 300 iterations with 10 iterations warm-up. In Fig.~\ref{fig:iou-flops} and Fig.~\ref{fig:iou-fps}, the segmentation quality (mIoU) of the models in relation to their computational complexity (FLOPs) and inference speed (FPS) are shown, respectively; and Fig.~\ref{fig:kyoto-tyrolw-maps} presents the visual comparison of their segmentation maps. The correlation between the efficiency indicators (i.e., FPS, FLOPs, and Params) of the models is presented in Fig.~\ref{fig:fps-flops-param}. Finally, Fig.~\ref{fig:UDA_results_continet} and Fig.~\ref{fig:uda_decrase_example} present the continent-wise domain generalisation results of the models.

\addtolength{\tabcolsep}{-2pt}
\begin{table*}[t]
\captionsetup{width=0.98\linewidth}
\caption{Semantic segmentation results of the baseline U-Net series models and the representative efficient DNNs handcrafted and automated neural architecture search (NAS) models on the test set of the OpenEarthMap benchmark. The best score for each metric is in \textbf{bold} and the second-best is \underline{underlined}.}
\label{tab:segmentation_results}
\begin{threeparttable}
    \begin{center}
    \scalebox{1.0}{
    \begin{tabular}{c c c c c c c c c c c c c c}
        \hline \hline
        \multirow{2}{*}{Method} & \multirow{2}{*}{Backbone} & \multicolumn{8}{c}{IoU (\%)}  & {mIoU} & {Params} & {FLOPs} & {Speed} \\ 
        \cline{3-10}
        &   &	Bareland    &	Rangeland	&	Developed	&	Road	&	Tree	&	Water	&	Agriculture	&	Building & (\%) $\uparrow$ & (M) $\downarrow$ & (G) $\downarrow$ & (FPS) $\uparrow$ \\	
        \hline
        \multicolumn{13}{c}{\textit{Handcrafted models}} \\
        \hline
        U-Net 	&	EfficientNet-B0$^\ast$ 	&	41.89	&	55.65	&	52.61	&	60.31	&	70.11	&	79.45	&	72.07	&	76.99	&	63.63	&	6.30	&	41.55	&	33.9	\\
        U-Net 	&	EfficientNet-B1$^\ast$ 	&	37.61	&	54.77	&	\underline{53.19}	&	\underline{61.89}	&	\underline{71.70}	&	80.63	&	72.60	&	\bf{78.11}	&	63.81	&	8.80	&	41.66	& 28.3		\\
        U-Net 	&	MoibleNet	&	39.45	&	54.09	&	52.34	&	59.85	&	70.05	&	77.08	&	70.85	&	76.30	&	62.50	&	6.90	&	55.49	& 44.3		\\
        U-Net 	&	MoibleOne\_S1	&	40.82	&	55.81	&	51.56	&	61.59	&	71.39	&	76.46	&	72.42	&	77.20	&	63.41	&	9.60	&	80.20	& 32.0		\\
        SegFormer	&	Mit-B0	&	28.54	&	52.31	&	46.24	&	51.34	&	68.10	&	73.35	&	67.06	&	69.69	&	57.08	&	3.72	&	25.56	&	42.4	\\
        Segmenter	&	Tiny	&	25.39    &	42.20  &	35.32        &	28.26  &	58.91 &	61.65    &	55.20     &	53.63  &	45.07 & 	6.71	&	18.24	&	22.4	\\
        HRNet	&	W18	    &	32.98     &	53.55     &	  48.17     &	57.45   &	69.41  &	75.48   &	70.01   &	73.28    &	60.04	&	9.68	&	76.32	&	14.8	\\
        STDC &	-    &	34.79     &	  53.28     &	  48.21     &	57.91     &	  68.99     &	79.38     &	70.09     &	71.99     &	    60.58   & 	8.53	&	35.12	&	78.3	\\
        CGNet &	-   &	35.06     &	  52.41  &	46.41      &	55.43     &	    68.16     &	78.08     &	69.67     &	71.63     &	59.61     & 	\underline{0.50}	&	13.72	&	56.7	\\
        FastSCNN &  -	&	34.75   &	53.27    &	48.31     &	   55.62  &	   68.83     &	78.94     &	70.43     &	71.90     &	60.26     & 	1.46	&	\bf{3.68}	&  \underline{110.1}  \\
        U-NetFormer 	&	EfficientNet-B0$^\ast$  &	41.88	&	55.17	&	52.69	&	60.89	&	70.92	&	82.03	&	\underline{73.78}	&	76.72	&	\underline{64.26}	&	4.11	&	15.78	&	13.1	\\
        U-NetFormer 	&	EfficientNet-B1$^\ast$ 	&	40.39	&	55.61	&	52.29	&	61.61	&	70.84	&	\bf{82.85}	&	73.15	&	77.22	&	\underline{64.24}	&	6.62	&	19.61	&	35.3	\\
        U-NetFormer 	&	EfficientNet-B2$^\ast$ 	&	39.17	&	\underline{56.07}	&	\bf{53.35}	&	\bf{62.10}	&	71.25	&	\underline{82.43}	&	\bf{74.28}	&	\underline{77.83}	&	\bf{64.56}	&	8.91	&	31.21	&	33.2	\\
        \hline
        \multicolumn{13}{c}{\textit{NAS models}} \\
        \hline
        MRF-UNets &  -	&	40.14	&	\bf{56.45}	&	52.95	&	61.27	&	\bf{71.79}	&	81.25	&	72.42	&	77.77	&	\underline{64.26}	&	1.62	&	38.71	&	21.4	\\
        BiX-NAS &   -	&	28.26	&	51.98	&	45.71	&	52.91	&	67.46	&	72.90	&	68.68	&	70.83	&	57.34	&	\bf{0.38}	&	112.29	&	29.3	\\
        SqueezeNAS &    -	&	40.90	&	55.35	&	50.27	&	58.70	&	69.88	&	82.08	&	73.75	&	74.49	&	63.18	&	1.82	&	11.36	&	81.6	\\
        DNAS &	-    &	\underline{43.19}	&	54.15	&	49.56	&	57.94	&	68.50	&	81.69	&	73.19	&	73.00	&	62.65	&	5.01	&	62.14	&	18.1	\\
        FasterSeg &	-   &	34.50	&	51.27	&	45.27	&	55.94	&	66.62	&	74.71	&	70.05	&	69.73	&	58.51	&	3.47	&	15.37	&	\bf{171.3}	\\
        SparseMask &    -	&	\bf{46.15}	&	51.88	&	44.01	&	43.64	&	65.20	&	77.41	&	71.48	&	66.11	&	58.23	&	2.96	&	\underline{10.28}	&	51.2	\\
        \hline \hline
    \end{tabular}}
    \begin{tablenotes}
       \item[*] {The base architectures of these backbones were originally designed via a neural architecture search approach~\cite{tan2019efficientnet}.} 
    \end{tablenotes}
    \vspace{-1mm}
    \end{center}
\end{threeparttable}
\end{table*}

\subsubsection{Quality Comparison}
\begin{figure}
\begin{center}
\includegraphics[width=0.95\linewidth]{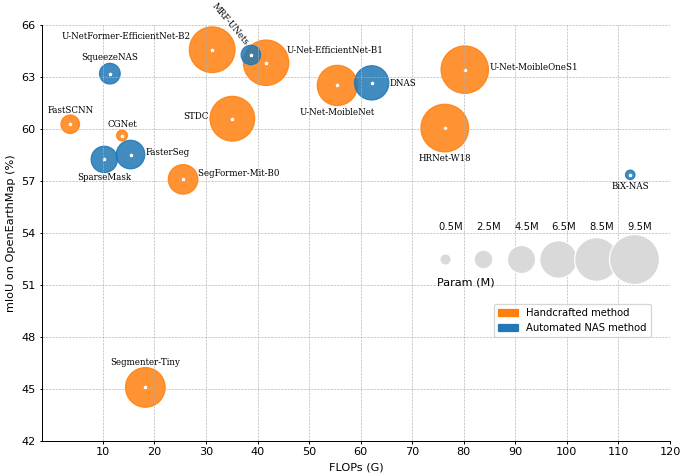}
\end{center}
\captionsetup{width=0.95\linewidth}
\caption{Segmentation quality (i.e., the mIoU on the OpenEarthMap dataset) versus computational complexity (FLOPs) of each model. The bubble size denotes the number of learnable parameters of the models, which literally indicates the size of the models.}
\label{fig:iou-flops}
\vspace{-2mm}
\end{figure}

\begin{figure}
\begin{center}
\includegraphics[width=0.95\linewidth]{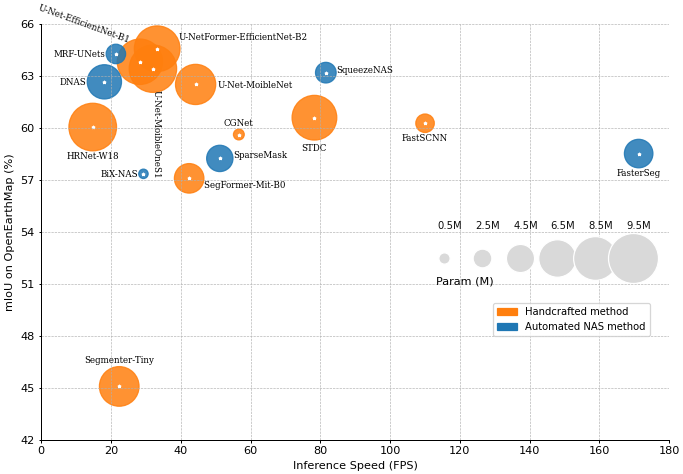}
\end{center}
\captionsetup{width=0.95\linewidth}
\caption{Segmentation quality (i.e., the mIoU on the OpenEarthMap dataset) versus inference speed (FPS) of each model. The bubble size denotes the number of learnable parameters of the models, which literally indicates the size of the models.}
\label{fig:iou-fps}
\vspace{-2mm}
\end{figure}

As shown in Table~\ref{tab:segmentation_results}, and in Fig.~\ref{fig:iou-flops} and Fig.~\ref{fig:iou-fps}, for real-time semantic segmentation in remote sensing applications, the existing handcrafted and automated architecture-searched compact networks could be considered the most practical starting points, as most of the efficient deep neural networks we evaluated did perform well on the OpenEarthMap test set. Apart from Segmenter-Tiny, as shown in Fig.~\ref{fig:iou-flops} and Fig.~\ref{fig:iou-fps}, the rest of the models achieved more than 50\% mIoU accuracy rate. Among these networks, the U-NetFormer-EfficientNet-B2 stands out in the overall segmentation quality with an accuracy rate of 64.56\% mIoU with 8.91M parameters. This may be attributed to the model's integration of global and local information using a global-local Transformer block in the decoder and leveraging the EfficientNet-B2 encoder to extract features.  Our findings suggest that combining a U-Net of Vision Transformer with an EfficientNet backbone is effective in generating local attention and increasing the effective receptive field size, ultimately leading to improved segmentation quality. 

\begin{figure*}
\vspace*{-2.5mm}
\captionsetup[subfloat]{captionskip=-2mm}
\centering
    \subfloat[]{\includegraphics[width=0.95\linewidth]{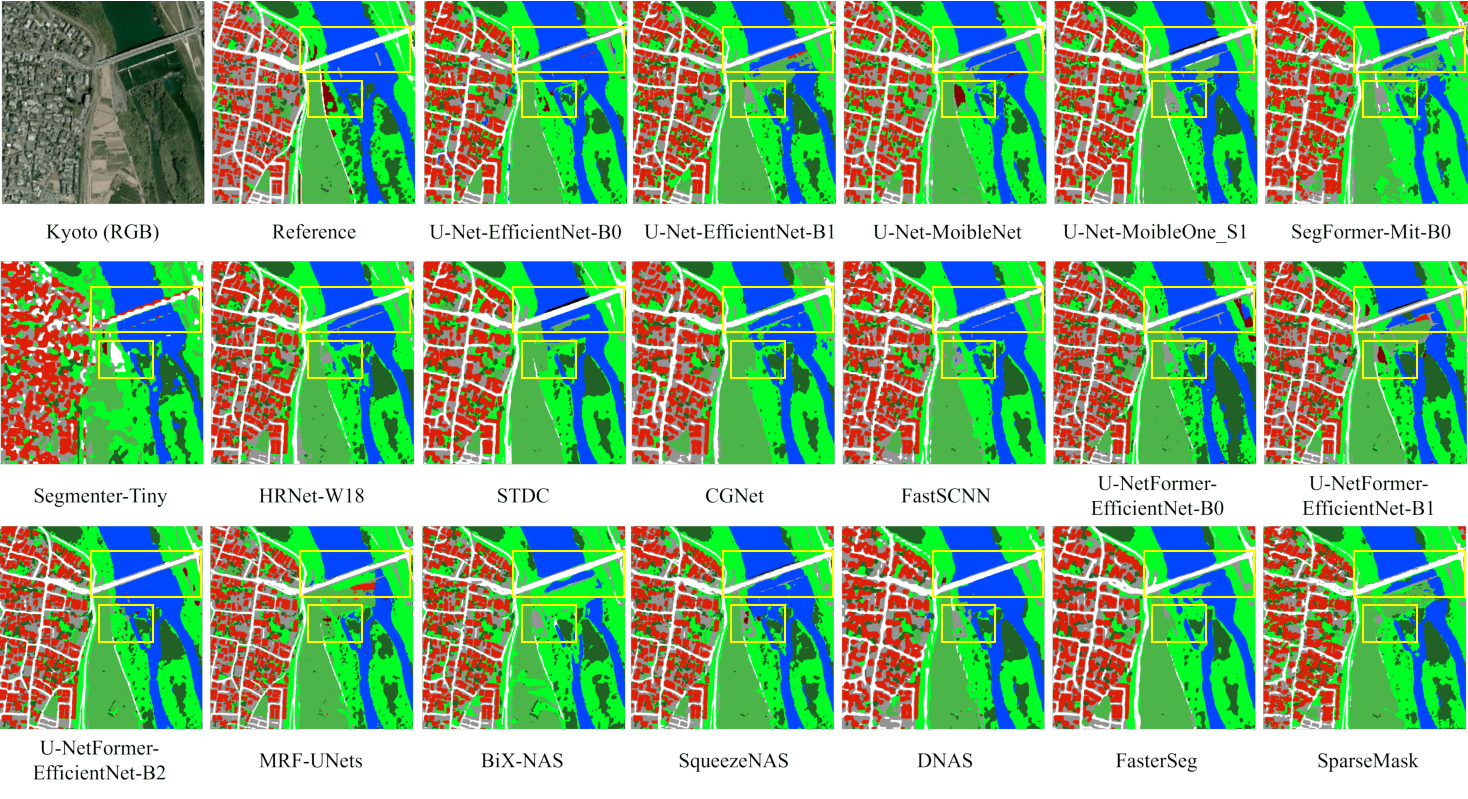}}\\
    \vspace{-1mm}
    \subfloat[]{\includegraphics[width=0.95\linewidth]{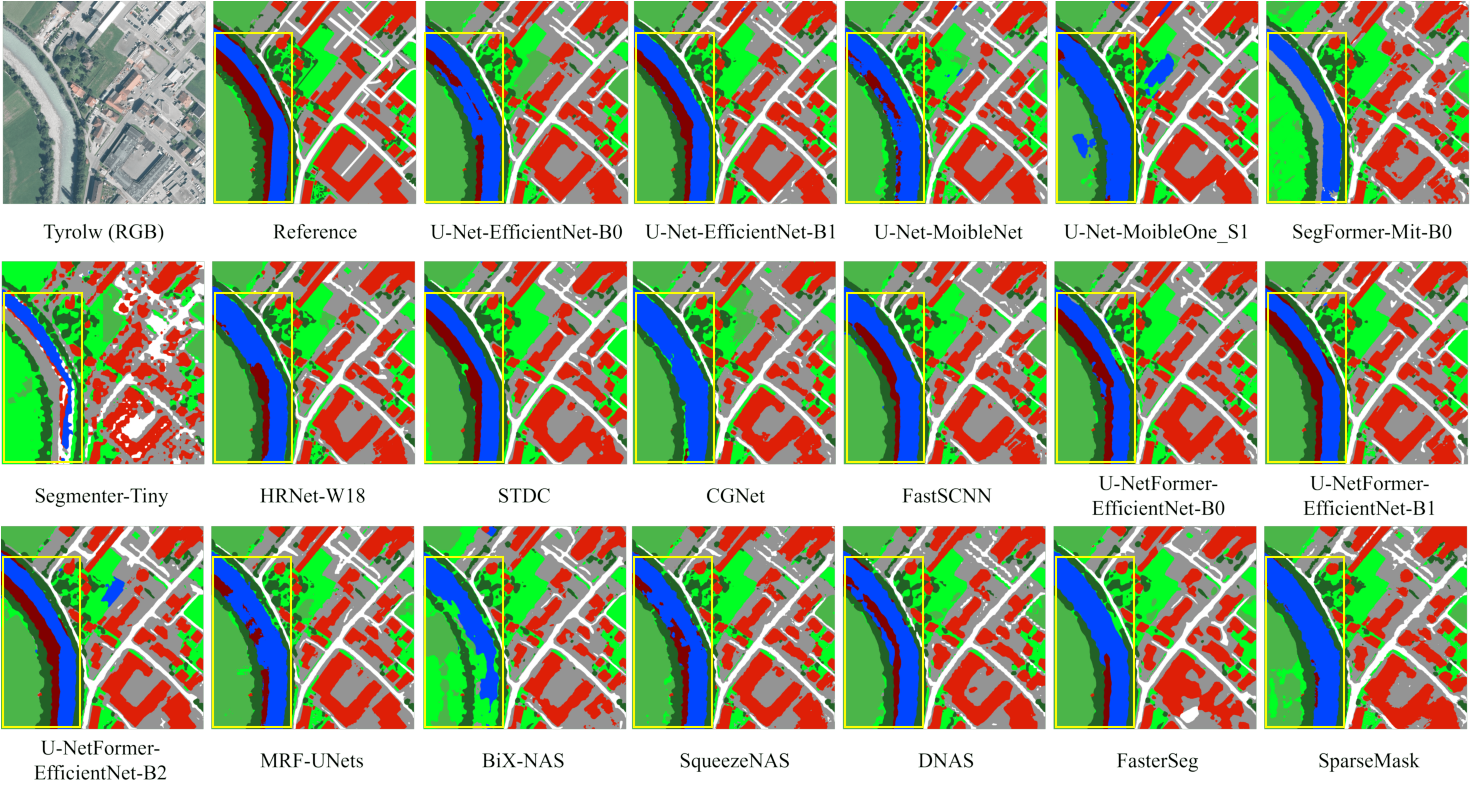}}\\
    \captionsetup{width=0.95\linewidth}
    \vspace{-1mm}
\caption{Visual comparison of land cover mapping results among the models on (a) Kyoto\_47 and (b) Tyrolw\_9 image patches of the OpenEarthMap benchmark test set. The highlighted regions (yellow rectangles) indicate the regions with significant differences in the maps produced by the models with respect to the reference map.}
\label{fig:kyoto-tyrolw-maps}
\vspace{-4mm}
\end{figure*}

The U-Net-based efficient networks, U-Net-EfficientNet-B0 and U-Net-EfficientNet-B1, which are commonly used for real-time remote sensing applications achieved an accuracy rate of 63.63\% mIoU and 63.81\% mIoU with 6.3M and 8.8M parameters, respectively. When the EfficientNet-B1 backbone is replaced with MobileNetV2 lightweight architecture, as demonstrated in U-Net-MobileNet, the number of parameters is reduced to 6.9M but with a slight sacrifice, i.e., a decrease of approx. 1.1\%, in the segmentation performance. This makes it a suitable option for applications in a low-power environment. To improve segmentation performance, we can slightly increase the number of trainable parameters of the backbone architecture. For example, increasing the number of parameters of the backbone to 9.6M by replacing MobileNetV2 with an enhanced version, MobileOne\_S1, improved the segmentation performance by about 1\%, i.e., from 62.50\% mIoU to 63.41\% mIoU. However, using 29\% fewer parameters, the U-NetFormer-EfficientNet-B0 and U-NetFormer-EfficientNet-B1, achieved better accuracy (approx. 1\% higher) than their U-Net-EfficientNet counterparts. This suggests that using U-NetFormer with EfficientNet lightweight backbone is more suitable for real-time semantic segmentation in remote sensing compared to the conventional U-Net with EfficientNet lightweight backbone. Moreover, as clearly shown in Fig.~\ref{fig:iou-flops} and Fig.~\ref{fig:iou-fps}, all the U-Net-based compact methods we studied outperformed the other handcrafted compact methods in terms of segmentation quality. The Segmenter-Tiny, the SegFormer-Mit-B0, and the CGNet yield accuracy rate below 60\% mIoU, with the results of 45.07\%, 57.08\%, and 59.61\% mIoUs, respectively. Other handcrafted compact methods, such as HRNet-W18, STDC, and FastSCNN achieved similar accuracy rates. FastSCNN maintains efficiency in terms of FLOPs and FPS and achieved an acceptable mIoU of 60.26\%. 



Most of the automated architecture search methods reduce the parameters of the network by compromising accuracy. For instance, BiX-NAS, FasterSeg and SparseMask, respectively with mIoUs of 57.34\%, 58.51\%, and 58.23\% are among the least accurate methods. However, with approximately 82\% fewer parameters, the MRF-UNets achieved 64.26\% mIoU to compete with the best handcrafted compact methods, the U-NetFormer-EfficientNet series (see Table~\ref{tab:segmentation_results}). This can clearly be shown in Fig.~\ref{fig:iou-flops} and Fig.~\ref{fig:iou-fps} as well. Furthermore, as shown in Table~\ref{tab:segmentation_results}, the SqueezeNAS and the DNAS also use 81\% and 27\% fewer parameters, respectively, to achieve segmentation accuracy rates of 63.18\% and 62.65\% mIoUs, which compete with U-Net-EfficientNet and U-Net-Mobile series of the handcrafted compact methods. Although the NAS methods trailed behind (slightly in some cases) the handcrafted ones in terms of segmentation quality, they offer the benefit of reducing the storage size of the network. To summarize, the results of the semantic segmentation experiments on the OpenEarthMap remote sensing benchmark indicate that the U-NetFormer handcrafted architectures with EfficientNet lightweight backbones may serve as the best baselines for the OpenEarthMap, but comparatively with a large number of parameters and FLOPs (see Fig.~\ref{fig:iou-flops}) as well as low inference speed (see Fig.~\ref{fig:iou-fps}). In contrast, the architectures based on the NAS approach offer reduced-scale models with fewer FLOPs, however, they exhibit some sacrifice in segmentation quality.

In addition to the quantitative (mIoU) evaluation of the quality of the models' segmentation results (see Table~\ref{tab:segmentation_results}),  Fig.~\ref{fig:kyoto-tyrolw-maps} provides a visual comparison of the segmentation maps of the models. The representative areas are  Kyoto and Tyrolw image patches of the OpenEarthMap benchmark test set. The U-NetFormer-EfficientNet-B2 produces the best-detailed visualization results. As shown in Fig.~\ref{fig:kyoto-tyrolw-maps}(a), the \textit{water} area of the dam was misclassified as \textit{rangeland} and \textit{agricultural land} by U-Net-EfficientNet-B1, U-Net-MoibleOne\_S1, SegFormer-Mit-B0, U-NetFormer-EfficientNet-B1, MRF-UNets, BiX-NAS, FasterSeg and SparseMask, while the other methods correctly identified them. Only the U-Net-MoibleNet and the U-NetFormer-EfficientNet-B1 are able to identify the \textit{bareland} nearby the river, while the other methods wrongly classified it as \textit{development area}. With the exception of Segmenter-Tiny, the other methods can produce well-maintained boundaries of the roads and buildings.

In Fig.~\ref{fig:kyoto-tyrolw-maps}(b),  the U-Net series and MRF-UNets can identify the tiny roads in the top-right parts of the image. The U-Net-EfficientNet-B1 and the U-NetFormer series recognize the \textit{bareland} along the rivers, while the other methods classified the area as \textit{water} and \textit{developed space}. Visually, the accurate maps produced by the models contain tremendous structured characteristics like rivers (water), buildings, and agriculture. The \textit{water} and \textit{bareland} classes which score the highest and lowest accuracy rates, respectively, across all the models (see Table~\ref{tab:segmentation_results}) is clearly demonstrated in the segmentation maps of the models. With the fragmented layouts and the different sizes, the models find it difficult to adequately define the limits of the roadways and the buildings. Because of the striking similarities in their spectra, \textit{rangeland}, \textit{agricultural land} and \textit{trees} sometimes confused the models. And since parking lots and cover materials in certain rural regions are comparable, \textit{roads} are frequently misclassified as \textit{developed space}.

\subsubsection{Efficiency Comparison}
In Table~\ref{tab:segmentation_results}, we also presented three efficiency indicators, that is, the number of parameters (Params), computational complexity (FLOPs), and inference time (FPS) to measure the models' efficiency with respect to storage (memory), computation, and latency, respectively. Fig.~\ref{fig:iou-flops} and Fig.~\ref{fig:iou-fps} also, respectively, provide FLOPs versus mIoU (segmentation quality) and FPS versus mIoU with respect to the Params of the models. For U-Net-EfficientNet and U-NetFormer-EfficientNet series, we used the U-Net-EfficientNet-B1 and the U-Net-EfficientNet-B2, respectively, for the efficiency comparison. 

As shown in Fig.~\ref{fig:iou-flops} and Fig.~\ref{fig:iou-fps}, most of the NAS compact models have fewer parameters compared to their handcrafted counterparts. This makes the NAS methods more efficient than the handcrafted compact methods in terms of storage (few parameters). For example, with 0.38$M$ parameters stored in 32-bit width (4 bytes), the Bix-NAS needs only 1.5MB of memory. However, some handcrafted compact networks are efficient in terms of storage. The CGNet handcrafted model needs only 2.0MB of memory to store its 0.5$M$ parameters in 32-bit width. The storage range for the NAS models is 1.5MB--20.1MB and for the handcrafted ones is 2.0MB--32.7MB based on 32-bit width. It must be noted that some of the models sacrificed segmentation quality for efficient storage. For example, CGNet, Bix-NAS, FasterSeg, SparseMask, and SegFormer achieved $<60\%$ mIoUs with storage ranging from 1.5MB to 14.8MB. However, the MRF-UNets NAS model did well in a quality-storage trade-off, achieving 64.26\% mIoU with only 6.5MB storage. Whereas, the U-Net-EfficientNet-B2 handcrafted model performed badly in a quality-storage trade-off, using a large storage of 35.6MB to achieve an accuracy rate of 64.56\% mIoU, equivalent to that of MRF-UNets (64.26\%) which needs only 6.5MB of memory. Hence, for a quality-storage trade-off, the MRF-UNets model is more desirable for real-time semantic segmentation in remote sensing applications.

\begin{figure}
\begin{center}
\includegraphics[width=0.95\linewidth]{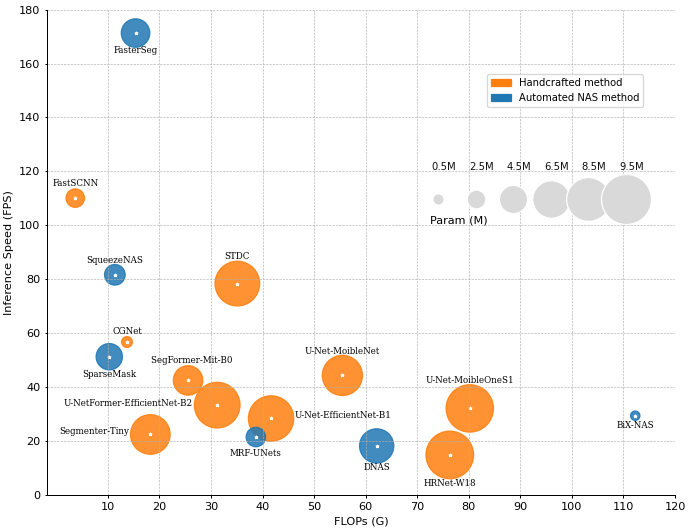}
\end{center}
\captionsetup{width=0.95\linewidth}
\caption{Correlation between the efficiency indicators of the models: inference speed (FPS) versus computational complexity (FLOPs) versus the number of learnable parameters (Params). The bubble size denotes the number of learnable parameters of the models, which literally indicates the size of the models.}
\label{fig:fps-flops-param}
\vspace{-2mm}
\end{figure}

On computation, it can be seen in Fig.~\ref{fig:iou-flops} that the U-Net-based and the U-NetFormer-based handcrafted models comparatively have high computational complexity (FLOPs). These models performed well in segmentation quality ($>62\%$ mIoUs) at the cost of high computation. SegFormer-Mit-B0, CGNet, and Segmenter-Tiny handcrafted models and NAS models like SparseMask and FasterSeg sacrificed segmentation quality ($<60\%$ mIoUs) for low-cost computation ($<25G$ FLOPs). The SqueezeNAS did best in the quality-computation trade-off, using only $11.36G$ FLOPs to achieve $<63.18\%$ mIoUs. The FastCNN handcrafted model has a fair quality-computation trade-off balance (achieving 60.26\% mIoU with $3.68G$ FLOPs). With $112.29G$ FLOPs, BiX-NAS achieved 57.34\% mIoU, which makes it the worst model in terms of quality-computation trade-off. The MRF-UNets (achieving 64.26\% mIoU with $38.71G$ FLOPs) trailed behind the U-NetFormer-EfficientNet-B2 (achieving 64.56\% mIoU with $8.91G$ FLOPs) in a quality-computation trade-off. This makes U-NetFormer-EfficientNet-B2 more attractive for real-time semantic segmentation in remote sensing with respect to a quality-computation trade-off.

In Fig.~\ref{fig:iou-fps}, we can see that most of the models we evaluated have low inference speed (FPS), hence, high latency. The FasterSeg has the highest inference speed (171.3 FPS), which is desirable for real-time semantic segmentation in remote sensing, but sacrifices segmentation quality (58.51\% mIoU). FastCNN ranks second in speed but it has a better quality-speed trade-off (achieving 60.26\% mIoU with 110.1 FPS) compared to FasterSeg (achieving 58.51\% mIoU with 171.3 FPS). The U-Net-based and the U-NetFormer-based models achieved better segmentation accuracy rates ($>62\%$ mIoUs) by sacrificing inference speed ($<44.3$ FPS). However, here too, the MRF-UNets (achieving 64.26\% mIoU with 21.4 FPS) trailed behind the U-NetFormer-EfficientNet-B2 (achieving 64.56\% mIoU with 33.2 FPS) in a quality-speed trade-off. The SqueezeNAS has a fair quality-speed trade-off balance (achieving 63.18\% mIoU with 81.6 FPS), which may be a good choice for real-time semantic segmentation in remote sensing applications. The worst model of quality-speed trade-off is the Segmenter-Tiny (achieving the lowest accuracy rate of 45.07\% mIoU with 22.4 FPS).

Considering the overall quality-efficiency trade-off, U-NetFormer-EfficientNet-B2 (achieving 64.56\% mIoU with $8.91M$ Params, $31.21G$ FLOPs, and 33.2 FPS), MRF-UNets (achieving 64.26\% mIoU with $1.62M$ Params, $38.71G$ FLOPs, and 21.4 FPS), and SqueezeNAS (achieving 63.18\% mIoU with $1.82M$ Params, $11.36G$ FLOPs, and 81.6 FPS), the SqueezeNAS would be recommended for real-time semantic segmentation application in remote sensing with a little bit sacrifice of quality. These findings demonstrate the trade-off between quality and efficiency in selecting real-time semantic segmentation methods for remote sensing applications. 

\begin{figure*}
\begin{center}
\includegraphics[width=0.95\linewidth]{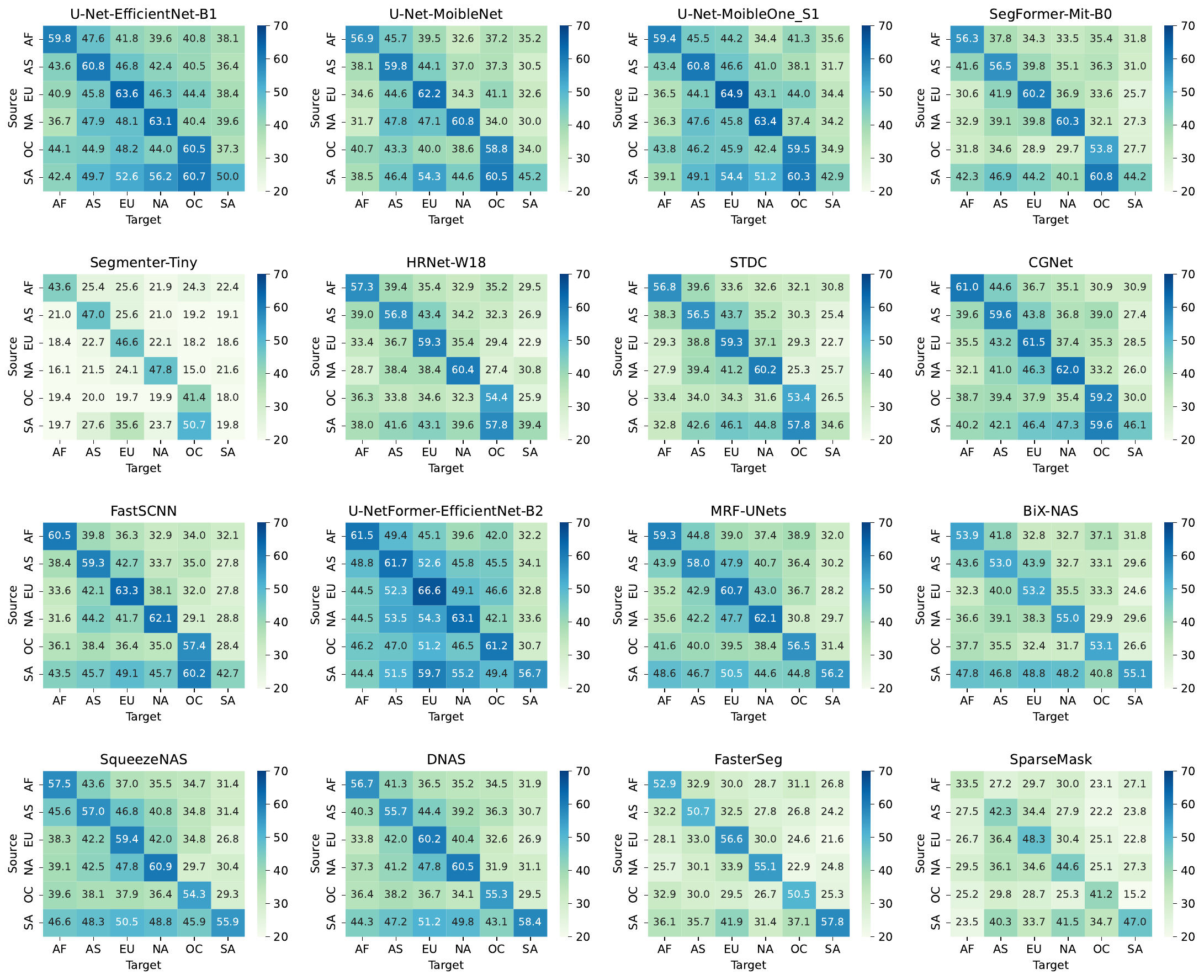}
\end{center}
\vspace{-4mm}
\captionsetup{width=0.9\linewidth}
\caption{Continent-wise domain generalisation results of the models. The results are mIoUs of the models based on the continent-wise domain adaptation settings in \cite{Xia2023WACV}. Asia: AS, Europe: EU, Africa: AF, North America: NA, South America: SA, and Oceania: OC. Note, for the U-Net-EfficientNet and the U-NetFormer-EfficientNet series, we only used U-Net-EfficientNet-B1 and U-Net-EfficientNet-B2, respectively, for this evaluation.}
\label{fig:UDA_results_continet}
\vspace{-4mm}
\end{figure*}

\begin{figure*}
\begin{center}
\includegraphics[width=0.95\linewidth]{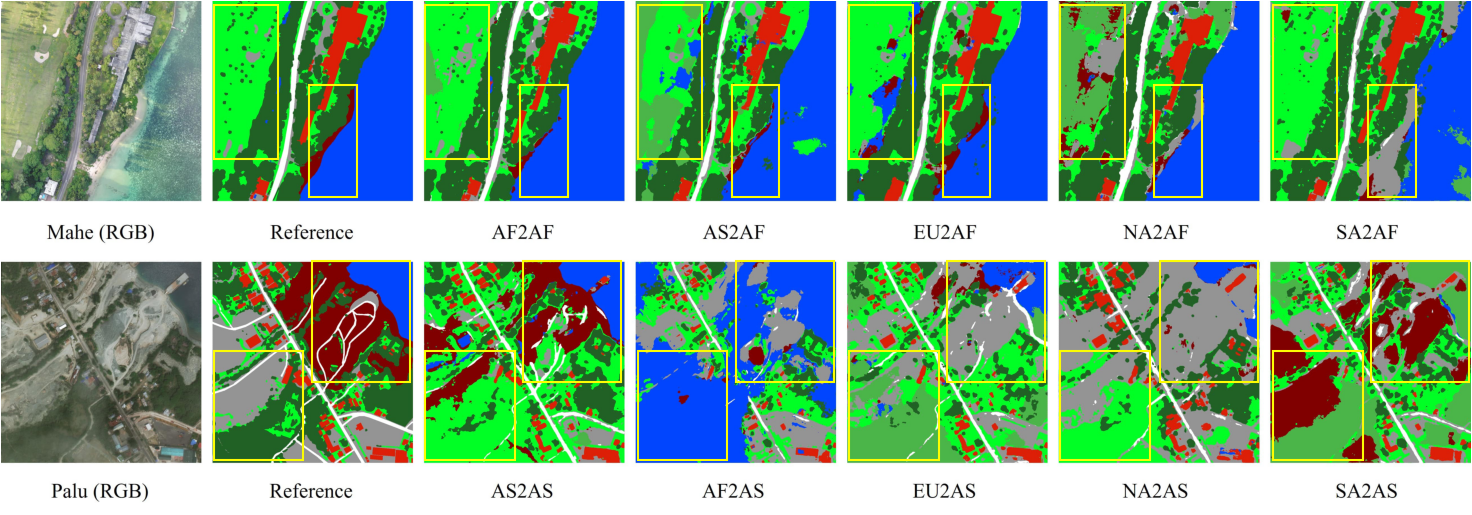}
\end{center}
\vspace{-3mm}
\captionsetup{width=0.95\linewidth}
\caption{Visual comparisons of continent-wise generalisation results of U-NetFormer-EfficientNet-B2 model. Asia: AS, Europe: EU, Africa: AF, North America: NA, South America: SA. The Mahe is from Africa (AF2AF performs best) and the Palu is from Asia (AS2AS performs best). AS2AF denotes AS as the source domain and AF as the target domain. The highlighted regions (yellow rectangles) indicate the regions with significant differences in the segmentation maps produced by the models with respect to the reference map.}
\label{fig:uda_decrase_example}
\vspace{-2mm}
\end{figure*}

\subsubsection{Correlation Between Efficiency Metrics}
It is commonly assumed that the efficiency indicators of the models are correlated (e.g., fewer parameters translate to lower computational complexity or higher inference speed). As shown in Fig.~\ref{fig:fps-flops-param}, this assumption might not necessarily be true, which is also observed by Cai \textit{et al.} \cite{cai2020}. For example, among all the models we evaluated, the BiX-NAS is the smallest with only $0.38M$ parameters; however, it has the largest FLOPs ($112.29G$) and low inference speed (29.3 FPS). Whereas STDC has more parameters ($8.53M$) than CGNet ($0.5M$), SparseMask ($2.96M$), and MRF-UNets ($1.62M$), the inference speed of STDC (78.3 FPS) is higher than that of CGNet (56.7 FPS), SparseMask (51.2 FPS), and MRF-UNets (21.4 FPS). Also, the FasterSeg which has the highest inference speed (171.3 FPS), has $3.09M$ and $2.97M$ more parameters than BiX-NAS ($0.38M$) and CGNet ($0.5M$), respectively. Furthermore, as shown in Fig.~\ref{fig:fps-flops-param}, most of the models we evaluated have a comparatively smaller number of FLOPs but with lower inference speeds. This also demonstrates that smaller FLOPs do not necessarily translate to higher inference speed. We found that models with architectures of more multi-branch connections tend to have a lower inference speed (e.g., SparseMask, MFR-UNets, HRNet, and BiX-NAS). These findings suggest that in selecting real-time semantic segmentation methods for remote sensing applications, it is not sufficient to rely on a single efficiency indicator as this can be misleading. Moreover, network architecture with minimal multi-branches should be adopted for high inference speed (i.e., low latency) in real-time applications of remote sensing semantic segmentation.

\subsubsection{Continent-Wise Generalisation}
Here, we investigated the continent-wise domain gap on the OpenEarthMap dataset. Fig.~\ref{fig:UDA_results_continet} presents continent-wise generalisation semantic segmentation results of the models we evaluated. The results are mIoUs of the models based on the continent-wise domain adaptation settings in \cite{Xia2023WACV}. For the U-Net-EfficientNet and the U-NetFormer-EfficientNet series, we used U-Net-EfficientNet-B1 and U-Net-EfficientNet-B2, respectively, for the evaluation. In general, the U-Net-based handcrafted models obtain better results than other methods in most cases, especially the U-NetFormer-EfficientNet-B2, which is quite similar to the results in the traditional semantic segmentation shown in Table~\ref{tab:segmentation_results}. The NAS models performed significantly lower ($<40\%$ mIoU) in most cases of the continent-wise domain generalisation settings. Even the MRF-UNets and the SqueezeNAS obtained comparatively worse results than the U-Net series, which differs from traditional semantic segmentation settings. The limited data of the Oceania (OC) continent led to the lowest transfer generalisation results when OC is treated as the source domain. In contrast, the performance with OC as the target domain is better than in settings of other continents. With the exception of the OC continent, most of the methods indicated two minor domain gaps: Europe-to-North America (EU2NA) and Asia-to-North America (AS2NA). The most prominent domain gaps the real-time semantic segmentation models reveal are Africa-to-Europe (AF2EU) and North America-to-Africa (NA2AF).

In Fig.~\ref{fig:uda_decrase_example}, we present two examples of the continent-wise domain gap based on the U-NetFormer-EfficientNet-B2 continent-wise domain generalisation segmentation map. When the Africa (AF) continent is used as the target domain (e.g., Mahe) and the other continents are used as the source domains, the segmentation quality significantly decreases compared to when the African continent is used as both source and target domains. In most cases, the \textit{rangeland} is misclassified as \textit{agricultural land} and \textit{water}. And some part of the ocean is wrongly classified as \textit{rangeland} and \textit{agricultural land}. Also, as shown in Fig.~\ref{fig:uda_decrase_example}, using Asia (AS) continent as the target domain (e.g., Palu) and the other continents as the source domains, Africa-to-Asia (AF2AS), Europe-to-Asia (EU2AS), and North America-to-Asia (NA2AS) produced bad segmentation in \textit{trees} and \textit{bareland}, and wrongly classified them as \textit{rangeland}, \textit{developed space} and \textit{agricultural land}. The \textit{bareland} and  \textit{water} in SA2AS are wrong identified to \textit{developed space} and \textit{agricultural land}. We suggest a combination of state-of-the-art unsupervised domain adaptation (UDA) techniques with efficient semantic segmentation methods to develop advanced compact UDA models suitable for continent-wise domain generalisation and continent-wise UDA tasks.
 
\vspace{-2mm}
\section{Conclusion and Open Challenges}\label{sec:6}
In this study, we discussed current developments in the literature regarding real-time semantic segmentation methods in remote sensing image analysis. We discussed network compression techniques and efficiency indicators of efficient deep neural networks for real-time semantic segmentation, and summarized the notable works proposed to address the problem of semantic segmentation in real-time applications of remote sensing on resource-constrained platforms. Furthermore, we performed extensive experiments with some existing real-time semantic segmentation methods, which include 13 handcrafted and 6 automated architecture-searched methods, on the OpenEarthMap remote sensing semantic segmentation benchmark to measure their quality-efficiency trade-off in real-world applications of remote sensing semantic segmentation. It appears to be the first time these models have been benchmarked on this particular dataset. We found that while generally there is a trade-off between segmentation quality and efficiency indicators (such as inference speed, computational complexity, and the number of parameters), most of the efficient deep neural networks we evaluated did achieve near state-of-the-art quality results but not with relatively fewer parameters capable of high inference speed. We also found that a model with very few parameters or less computational complexity is not necessarily fast during inference time. Overall, this study provides comprehensive insights into the developments in real-time semantic segmentation methods for remote sensing applications and highlights their strengths and weaknesses. The findings can inform future research in this field and help practitioners and researchers develop more efficient and accurate models for remote sensing applications. Here, we discuss some open challenges that can be adopted in future research efforts towards improving the quality-efficiency performance of real-time semantic segmentation deep learning methods in remote sensing image analysis.
\begin{itemize}
\renewcommand{\labelitemi}{\tiny$\blacksquare$}
    \item How to design a specialized efficient neural network backbone to improve the balance between accuracy and speed for accurate-fast segmentation of remotely-sensed images in real time? Exploring and adapting existing lightweight networks like MobileNet or EfficientNet via an automated architecture search approach for more suitable real-time segmentation methods in remote sensing may be an interesting challenge. 
    \item How to improve the domain generalisation performance as well as unsupervised domain adaptation of real-time semantic segmentation in remote sensing? The segmentation quality of the continent-wise domain generalisation is far behind that of the supervised semantic segmentation. Future work may adopt strategies in related tasks that have been proven successful such as self-training and curriculum learning to improve the accuracy of real-time continent-wise domain generalisation and continent-wise unsupervised domain adaptation.
    \item How to solve the variations in resolution and different sensor types problem of remote sensing imagery? Remote sensing imagery often comes from different sensors (optical or SAR) at different resolutions (meter or sub-meter), which can make it difficult for the algorithm to differentiate between different land cover classes. Exploring novel approaches to process images from different sensors at various resolutions in real-time segmentation represents an intriguing research direction in remote sensing image understanding.
    \item How to improve the robustness of real-time semantic segmentation algorithms to different imaging conditions such as weather conditions, seasonality, and temporal changes? Remote sensing images can be affected by various environmental factors, and developing efficient semantic segmentation networks that can adapt to these factors with a better balance between accuracy and speed in real time is an important challenge.
    \item How to improve the scalability of real-time semantic segmentation algorithms for large-scale remote sensing applications? Remote sensing images can cover vast areas, and processing such large datasets (e.g., county-wide or province-wide) in real-time could be a significant challenge. Developing efficient-scalable algorithms that can segment large-scale remote sensing data in real-time is essential for province-wide or county-wide remote sensing applications such as forest fire monitoring.
\end{itemize}


\bstctlcite{IEEEexample:BSTcontrol}
\bibliographystyle{IEEEtranN}
\bibliography{mybibfile}

\balance

\clearpage

\end{document}